%% file: neurips_2026.tex
\documentclass{article}
\usepackage{array}

 \usepackage[main, final]{neurips_2026}

\usepackage[utf8]{inputenc} 
\usepackage[T1]{fontenc}    
\usepackage{hyperref}       
\usepackage{url}            
\usepackage{booktabs}       
\usepackage{amsfonts}       
\usepackage{nicefrac}       
\usepackage{microtype}      
\usepackage[table]{xcolor}  
\usepackage{graphicx}
\usepackage{multirow}
\usepackage{subcaption}
\usepackage{caption}
\usepackage{float} 
\usepackage{algorithm}
\usepackage{algpseudocode}  
\usepackage{adjustbox}
\usepackage{pifont}
\usepackage{amsmath}
\usepackage{listings}
\usepackage[most]{tcolorbox}
\usepackage[normalem]{ulem} 
\usepackage{makecell}

\usepackage{tikz}
\usetikzlibrary{positioning, calc, arrows.meta}

\title{Complementing DINO Features with Image Structure for Part Discovery}

\author{%
  Samyak Rawlekar \\
  University of Illinois Urbana-Champaign\\
  Urbana, IL 61801\\
  \texttt{samyakr2@illinois.edu}
  \And
  Nikhil Chowdary Paleti  \\
   Independent Researcher \\
  Sunnyvale, CA, 94087\\
  \texttt{nikhilpaleti23@gmail.com}
  \And
  Amey Gupta \\
  University of Illinois Urbana-Champaign\\
  Urbana, IL 61801\\
  \texttt{amey2@illinois.edu}
  \And
  Narendra Ahuja \\
  University of Illinois Urbana-Champaign\\
  Urbana, IL 61801\\
  \texttt{n-ahuja@illinois.edu}
}

\begin{document}

\maketitle

\begin{abstract}
Self-supervised Vision Transformers (ViTs) like DINO encode rich semantics, enabling training-free segmentation of objects from the background and discovery of their parts via feature clustering. This paper focuses on part discovery: partitioning a foreground object into semantic parts. Existing approaches produce imprecise part boundaries. We trace this to the rigid ViT patch grid used in training DINO, where patches straddling boundaries contain pixels from multiple parts. While DINO's invariance-based training could absorb this pixel-level mixing into clean features, experiments with over 26K foreground patches show it does not: boundary-overlapping patches produce features with higher entropy and lower assignment confidence, blurring predicted part boundaries. 
Rather than make the massive effort of retraining DINO over geometrically coherent regions, an easier approach is to take boundaries directly from a boundary detector, which supplies the geometric information DINO's features lack.
Motivated by this, we introduce Part-DINO, a training-free framework that combines boundaries from low-level image segmentation with DINO's semantic features. We use the boundaries to partition the image into regions, pool DINO features within each region, and group them to discover parts. Without any training, Part-DINO produces structurally coherent part assignments, outperforming training-free baselines across four standard benchmarks (CUB, CelebA, PartImageNet-OOD, PartImageNet-Seg) and remaining competitive with weakly-supervised and unsupervised methods. 
\end{abstract}

\section{Introduction}
\label{sec:Intro}
Self-supervised Vision Transformers (ViTs) encode rich object-level structure. Models like DINO \citep{dino}, trained via self-distillation without any human labels, produce feature spaces that group semantic concepts. For example, on a bird, the eye patches cluster with eyes and wing patches with wings. This emergent property has enabled training-free object localization, segmentation, and coarse semantic part discovery. However, despite their strong semantic awareness, existing patch-based methods remain spatially imprecise, part assignments bleed across part contours, producing blurry boundaries between neighboring parts.

\begin{figure*}[t]
    \centering
    \resizebox{\linewidth}{!}{
        \begin{tabular}{c c c c} 
            \includegraphics[width=0.21\textwidth, height=3cm]{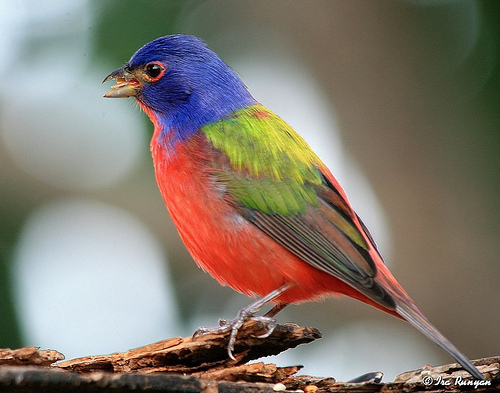} &
            
            \includegraphics[width=0.21\textwidth, height=3cm]{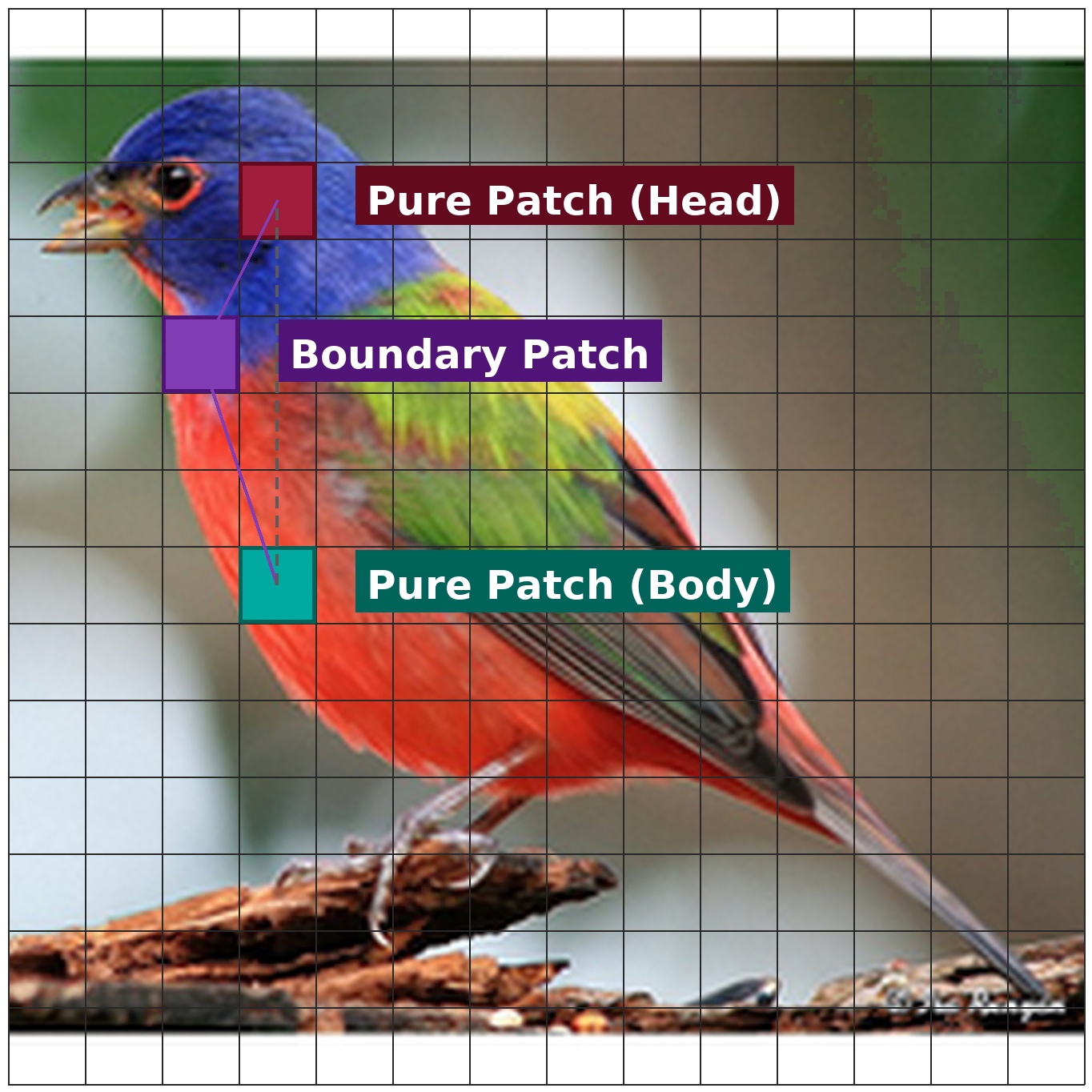} & 
            
            \includegraphics[width=0.27\textwidth, height=3cm]{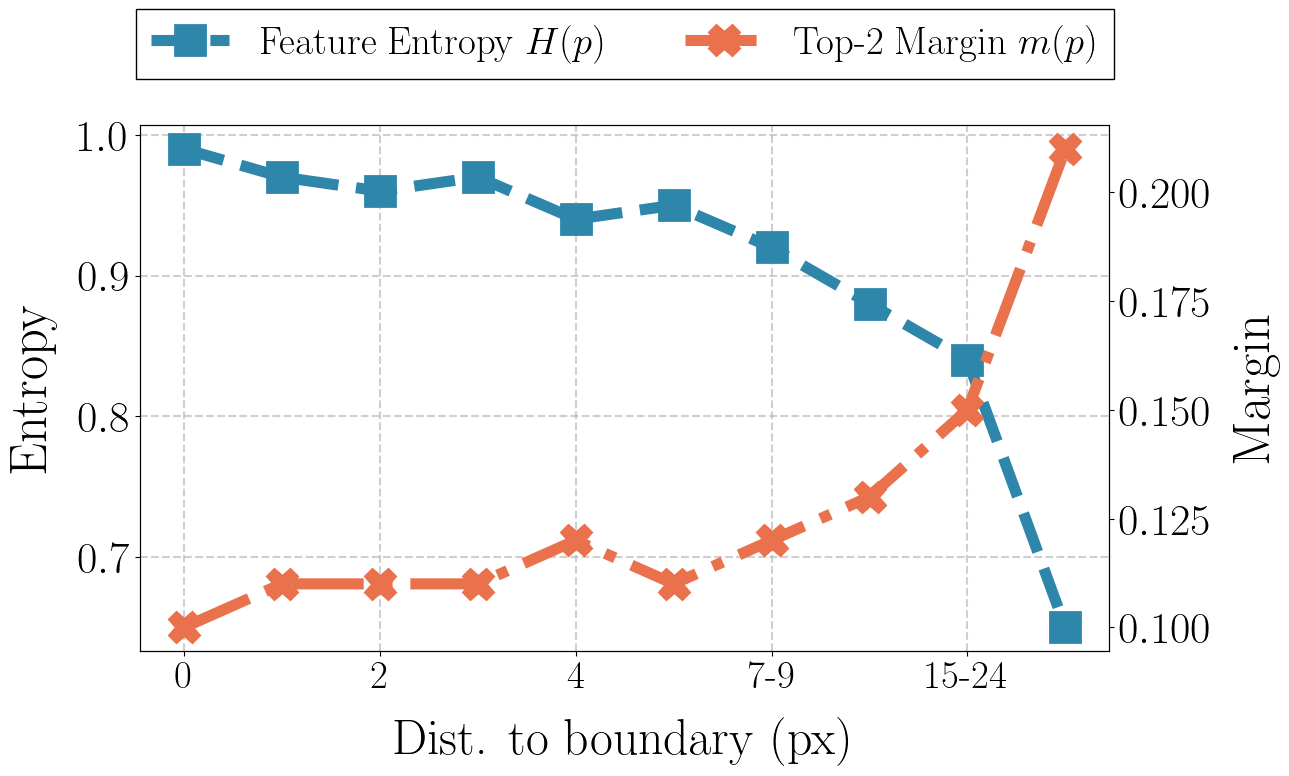} & 
            
            \begin{tabular}[b]{@{}c@{}}
                \textsf{\small DINO Similarity} \\[0.5ex]
                \includegraphics[width=0.2\textwidth, height=2.65cm]{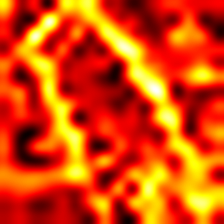}
            \end{tabular} \hspace{0.005\textwidth}
            \begin{tabular}[b]{@{}c@{}}
                \textsf{\small Image Contrast} \\[0.5ex]
                \includegraphics[width=0.2\textwidth, height=2.65cm]{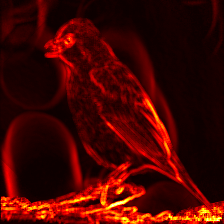}
            \end{tabular} \\
            
            (a) Input Image & (b) Geometric Mixing & (c) Feature Mixing & (d) Semantic vs. Structural Edges \\
        \end{tabular}
    } 
    \caption{\textbf{Mixing of pixels from across part boundaries present in boundary-area patches blurs semantic boundaries.} (a) Input image. (b) A boundary patch straddling the bird's head and body pools pixels from both regions, illustrating geometric mixing in pixel space. (c) Across 26K foreground patches, feature entropy rises and the top-2 assignment margin falls as patches approach part boundaries, showing mixing in feature space, which causes ambiguity in inferences of semantic part boundaries as can be seen in (d) where edges from DINO feature similarity (left) blur at part boundaries, while edges from image contrast (right) align with them. Recovery of sharp boundaries from DINO features may thus be expected to benefit from complementary geometric information.
    }
    \label{fig:teaser}
\end{figure*}

Why does a model with such semantic awareness fail to localize where one part ends and another begins? We trace this to the rigid patch grid over which features are computed. ViTs divide images into a fixed-size square patches that are agnostic to object geometry. When a patch straddles a part boundary, it pool pixels from multiple semantic regions before feature extraction (Fig~\ref{fig:teaser}b). While DINO's training encourages invariance across augmented views, it does not guarantee that this pixel-level mixing is resolved in the learned features. To investigate this, we analyze how patch features behave as a function of a patch's distance to the nearest ground-truth part boundary. Analyzing 26K foreground patches, we find that as patch–boundary overlap increases, feature entropy rises and assignment confidence falls (Fig~\ref{fig:teaser}c), confirming pixel-level mixing propagates into feature space.

This finding has a direct consequence: DINO's semantic information is useful for grouping parts, but unreliable for identifying their boundaries. As shown in Fig~\ref{fig:teaser}d, the edges derived from DINO feature similarity are blurred at part contours, whereas edges derived from image contrast are aligned with physical boundaries. Rather than retrain DINO over geometrically coherent regions at massive cost, an easier approach is to fuse it with an independent boundary detector.

Motivated by this, we propose Part-DINO, a training-free framework that combines the accurate boundaries from low-level image segmentation with the semantic information of DINO features. The boundaries partition the image into regions whose contours align with object geometry. We then pool DINO features within each region and use them for semantic grouping, assigning each region to its nearest part prototype.


We evaluate Part-DINO across four standard part-discovery benchmarks (CUB \citep{cub}, CelebA \citep{celevA}, and the OOD and Seg splits of PartImageNet\citep{Partimagenet}), using standard metrics (NME, NMI, and ARI). Part-DINO is training-free: it relies only on the frozen DINO features and avoids the optimization required by unsupervised methods (which train without labels) and weakly-supervised methods (which use image-level labels). Despite this constraint, Part-DINO outperforms the patch-based training-free baseline on NME and ARI across all datasets, and on NMI for three of four. The slight NMI drop on PartImageNet-OOD reflects a known metric trade-off, where ARI rewards spatial coherence over fine fragmentation (Sec.\ref{sec:experiments}). Furthermore, Part-DINO remains competitive with trained models, demonstrating that integrating image geometry with frozen semantic features recovers part structure without learned objectives. 

Our contributions in this paper are:

\begin{enumerate}
    \item We show that DINO's spatial imprecision stems from the ViT's patch grid. By analyzing features of patches that span boundaries, we find that pixel-level mixing propagates into the feature space, thereby blurring predicted contours.

    \item We introduce Part-DINO, a training-free framework that integrates the boundary of low-level image segments to extract sharp regions and DINO features to semantically group them, achieving precise part discovery without dataset-specific training.
    \item We evaluate Part-DINO across four diverse part-discovery benchmarks, demonstrating that it outperforms existing training-free baselines and is competitive with trained weakly-supervised and unsupervised methods, without any training.
\end{enumerate}


\section{Related Work}
\noindent\textbf{Part Discovery.} This task aims to identify semantically consistent region (parts) across varying instances of an object category. Earlier approaches rely on weak supervision, discovering parts by using image-level category labels to optimize a classification objective that highlights discriminative regions \citep{huang2020interpretable,pdisconet,pdiscoformer}. Another line of work explores unsupervised methods that do not rely on any part annotations. Some approaches train a masked autoencoder that masks image patches and learns to reconstruct them \citep{mpae}, while others use an equivariance-based objective \citep{xia2024unsupervised}. However, both weakly and unsupervised methods need a computationally expensive and dataset-specific training phase.  As an alternative, training-free approaches \citep{amir2021deep} bypass optimization entirely by clustering patch features from a frozen DINO backbone. Part-DINO follows this training-free paradigm, but use geometric regions as units instead of patches.

\noindent\textbf{Self-Supervised Vision Models.}
DINO~\citep{dino,dinov2} trains a Vision Transformer via self-distillation, producing patch-level embeddings that are consistent across augmented views and naturally group semantic concepts. These features encode strong spatial layouts and have been leveraged for foreground localization~\citep{simeoni2021localizing,tokencut,object-DINO} and coarse part discovery~\citep{amir2021deep}. Prior work has also investigated semantic entanglement in pretrained vision-language representations, demonstrating its impact on multi-object localization~\citep{dclip}. However, the unit used by these methods (the fixed-size square patch) is agnostic to object boundaries. Because this rigid grid ignores true object contours, boundary patches include multiple parts, producing semantically mixed features that limit spatial precision. Part-DINO avoids this by replacing the patch with geometric regions as the atomic unit, ensuring each feature vector corresponds to a single coherent part.

\noindent\textbf{Region Generation.} To avoid feature mixing, recent works group pixels into atomic units \citep{shlapentokh2024region}, often relying on superpixel algorithms like SLIC \citep{slic} to define these regions. However, while SLIC clusters pixels in a joint color-spatial space and partially respects image gradients, it imposes an explicit compactness constraint that trades boundary adherence for spatial uniformity. When part contours are weak relative to this compactness penalty, SLIC superpixels bleed across the irregular edges of natural object parts. Instead, we use the Ramp Transform and Watershed \citep{akbas_ahuja, akbas2014low}. By defining the units via contrast gradients (ramp edge discontinuities), we generate regions that adhere to low-level structural boundaries.

\begin{figure}
    \centering
    \includegraphics[width=0.9\linewidth]{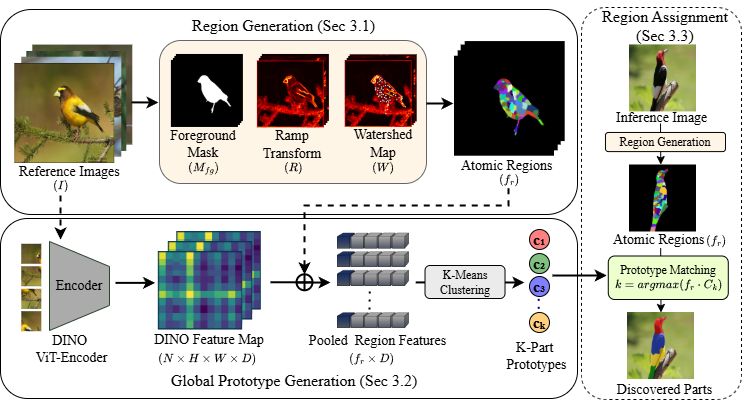}
\caption{\textbf{Overview of Part-DINO.} Given a set of reference images $\mathcal{I}$, the Region Generation module (Sec.\ref{sec:region_generation}, top) extracts a foreground mask of the object $M_{fg}$ and performs low-level image segmentation to produce atomic regions. In parallel, the Prototype Generation module (Sec.\ref{sec:prototype_generation}, bottom) encodes the reference images through a frozen DINO ViT to obtain dense feature maps. These semantic features are pooled within each atomic region and clustered via K-Means to establish $K$ global part prototypes $\{c_1, c_2, \dots, c_K\}$. At inference (right), the Region Assignment module (Sec.\ref{sec:region_assignment}) applies the same region generation pipeline to a test image and assigns each region to its nearest semantic prototype, producing the final part labels without any training.}
    \label{fig:method}
\end{figure}
\section{Part-DINO}
\label{sec:method_analysis_brief}
\textbf{The Boundary Patches:} 
We now detail the analysis behind the finding that pixel-space mixing propagates into feature space Sec.\ref{sec:Intro}. For each foreground patch $p$, we measure four quantities: (1) Euclidean distance $d(p)$ of its center to the nearest ground-truth part boundary, (2) its spatial purity $\pi(p) = |\text{majority-part pixels}|/|p_\text{fg}|$, (3) its feature entropy $H(p)$, computed as the entropy of the patch’s softmax similarity to per-part mean features extracted from clean interior patches, and (4) its top-2 assignment margin $m(p)$,  the similarity gap between $p$'s closest and second-closest part vector.
%
%

Across $26,459$ foreground patches, smaller $d$ corresponds to larger patch–boundary overlap. As overlap increases, it causes spatial purity ($\pi(p)$) to drop (Spearman correlation $\rho(d, \pi) = +0.71$, Supp. Fig.\ref{supp_fig:analysis}a), confirming pixel-space mixing. The key finding is that this pixel space mixing propagates into the feature space: patches closer to a boundary show higher feature entropy ($\rho(d, H) = -0.38$) and a lower top-2 assignment margin ($\rho(d, m) = +0.36$). Furthermore, this boundary mixing is widespread: $21.9\%$ of all foreground patches have a purity of $\pi<1$. As shown in Figure~\ref{fig:teaser}c, $H$ and $m$ vary smoothly with distance $d$, therefore the feature quality degrades continuously as overlap increases.

This analysis has two direct implications for part discovery. First, the spatial unit must respect object geometry, since a patch that spans multiple parts cannot preserve the contour separating them. Second, the geometric signal must be independent of DINO: as Fig.\ref{fig:teaser}d shows, edges derived from DINO feature similarity blur at part contours, whereas edges from image contrast remain sharp and aligned with physical boundaries. Part-DINO follows from these two constraints: we extract regions from a low-level segmentation of the image without relying on DINO's semantic features(Section~\ref{sec:region_generation}), build part prototypes from these regions (Section~\ref{sec:prototype_generation}), and assign part labels by matching region features to prototypes (Section~\ref{sec:region_assignment}).

\input{Algo/Extract_Region_Features}

\input{Algo/Global_Prototype_Generation}
\subsection{Region Generation}
\label{sec:region_generation}

Given an input image $I \in \mathbb{R}^{H \times W \times 3}$, our goal is to parse the object into a set of geometrically coherent regions $\mathcal{R}$ that respect part boundaries. Since we are interested in parts of the object rather than the background, we first isolate the foreground by extracting a binary mask $M_{fg} \in \{0,1\}^{H \times W}$ using foreground extraction methods. Within the foreground $M$, we compute a structural edge map $E_{\text{edge}}$ from the image contrast. We apply the RAMP transform~\cite{akbas_ahuja}, which estimates a per-pixel contrast magnitude based on local intensity transitions across four orientations  ($0^\circ, 45^\circ, 90^\circ, 135^\circ$). By construction, $E_{\text{edge}}$ depends only on pixel intensities, making it independent of DINO's feature pipeline that runs in parallel to extract semantic descriptors. 

To generate the atomic regions from this edge map, we apply the Watershed transform \cite{vincent1991watersheds} to $E_{\text{edge}}$, which treats the edge map as a topographic surface and floods upward from local minima, placing region boundaries along the ridges where adjacent basins meet. Seeds are placed at the local minima of $E_{\text{edge}}$ with a minimum inter-seed distance of $d_{\min}$ pixels. Watershed tends to over-segment near weak edges, producing spurious micro-regions that  lack meaningful structure.  We resolve this by merging any region under $s_{\min}$  pixels into its most similar neighbor (via DINO cosine similarity)  producing the final region set $\mathcal{R}$. Since merging only removes boundaries and never places one, every boundary within the object remains a watershed ridge of $E_{\text{edge}}$; setting $s_{\min}=0$ skips the merge entirely (Tab.~\ref{tab:merge}).

\subsection{Global Prototype Generation}
\label{sec:prototype_generation}

Building upon the regions $\mathcal{R}$ extracted in Section~\ref{sec:region_generation}, we compute a semantic representation for each such region $r \in \mathcal{R}$. We first extract the dense DINO feature map $F_{\text{patch}} \in \mathbb{R}^{H_p \times W_p \times D}$ from the input image, where $D$ is the embedding dimension and $(H_p, W_p)$ is the patch grid. We bilinearly upsample this to the image resolution $(H, W)$ to get a per-pixel feature map $F \in \mathbb{R}^{H \times W \times D}$ in which each pixel $p$ has a feature vector $F_p \in \mathbb{R}^D$. Each region's representation is then the $\ell_2$-normalised average of these features over the pixels assigned to that region:
\begin{equation}
\hat{f}_r = \frac{1}{|r|} \sum_{p \in r} \frac{F_p}{\|F_p\|_2}
\label{eq:region_feature}
\end{equation}
Because each region is bounded by high image contrast pixels rather than by the fixed contour of the predetermined patch grid, the pooled feature $\hat{f}_r$ aggregates pixels that lie within a single part, avoiding the loss of boundary information that compromises patch-level features.

To discover $K$ globally consistent parts of an object category, we randomly sample a reference subset $\mathcal{D}_{\text{ref}}$ of $N$ images from the dataset. We then perform $K$-Means clustering over the complete global pool of region features ({$\hat{f}_r$) from this subset to establish our category-specific semantic dictionary:
\begin{equation}
\mathcal{C} = {c_1, \dots, c_K} = K\text{-Means}\left(\bigcup_{I_i \in \mathcal{D}_{\text{ref}}} \mathcal{F}_i; K\right)
\label{eq:global_centroids}
\end{equation}
where $\mathcal{F}_i$ is the set of region features extracted from image $I_i$. Each centroid $c_k \in \mathcal{C}$ acts as a global prototype for a specific part, aggregated across instances in the reference set ($\mathcal{D}_{\text{ref}}$). For datasets containing multiple object categories, this dictionary is generated independently for each class.

\subsection{Region Assignment}
\label{sec:region_assignment}
Given a new image $I_{\text{test}}$, we extract its region set ($\mathcal{R})$ and the pooled features ({$\hat{f}_r$) following Section~\ref{sec:region_generation}. Each region $r \in \mathcal{R}$ is then assigned to its nearest global prototype by maximizing the cosine similarity:\begin{equation}k^* = \arg\max_{k \in \{1,\dots,K\}} \left( \hat{f}_r^\top c_k \right),
\label{eq:region_assignment}
\end{equation}
producing a dense part label map $L$. Since both the extracted region features and the prototypes $\mathcal{C}$ lie in the same DINO embedding space, this assignment requires no additional training. Every pixel within a region receives the same label by construction, so the final part boundaries in $L$ coincide with the watershed boundaries computed from $E_{\text{edge}}$, not the rigid, low-resolution patch grid.
 For datasets containing multiple object categories, we maintain a separate prototype dictionary $\mathcal{C}^{(s)}$ for each category $s \in \mathcal{S}$ (Section~\ref{sec:prototype_generation}). During assignment, we first identify the category of $I_{\text{test}}$ by extracting its CLS token $z_{\text{cls}}$ and compare it against per-category CLS prototypes:
\begin{equation}
s^* = \arg\max_{s \in \mathcal{S}} \left( z_{\text{cls}}^\top \, \bar{z}_{\text{cls}}^{(s)} \right), \quad \text{where } \bar{z}_{\text{cls}}^{(s)} = \frac{1}{|\mathcal{D}_{\text{ref}}^{(s)}|} \sum_{I \in \mathcal{D}_{\text{ref}}^{(s)}} \text{CLS}(I)
\label{eq:category}
\end{equation}
where both $z_{\text{cls}}$ and $\bar{z}_{\text{cls}}^{(s)}$ are $\ell_2$-normalized. Region assignment then proceeds using the selected dictionary $\mathcal{C}^{(s^*)}$. This step is itself training-free, relying solely on the CLS token's global image representation to identify the object category before part-level assignment begins.

\begin{table}
\small
\captionof{table}{Quantitative comparisons on CUB dataset across part granularities K $\in \{4, 8, 16\}$. We report NME ($\downarrow$), NMI ($\uparrow$), and ARI ($\uparrow$) against weakly-supervised, unsupervised, and training-free methods.$\dagger$ indicates replicated with SAM foreground mask}
\label{tab:cub_benchmark}
\centering
\resizebox{0.97\linewidth}{!}{
\input{tables/main_cub}
}
\end{table}

\begin{table}
\small
\captionof{table}{Quantitative comparison on CelebA across part granularities K $\in \{4, 8, 16\}$. We report NME ($\downarrow$), NMI ($\uparrow$), and ARI ($\uparrow$) against weakly-supervised, unsupervised, and training-free methods. $\dagger$  indicates replication with SAM foreground masks.
}
\label{tab:celeba_benchmark}
\centering
\resizebox{0.97\linewidth}{!}{
\input{tables/main_celebA}
}
\end{table}

\section{Experiments}
\label{sec:experiments}
\textbf{Dataset \& Implementation Details.} We evaluate Part-DINO on four widely-used benchmarks: CUB \citep{cub}, CelebA \citep{celevA}, PartImageNet Out-of-Distribution (OOD) and PartImageNet Segmentation \citep{Partimagenet}. Following prior works, we use DINOv2 ViT-B/14 with registers \citep{vis_reg} as our frozen feature encoder, and set the input resolution to $448 \times 448$ for CUB and $224 \times 224$ for the remaining datasets. Foreground masks $M$ are obtained using SAM3 \citep{sam3}. The RAMP transform is computed along four orientations ($0^\circ, 45^\circ, 90^\circ, 135^\circ$) with a maximum ramp width of 5 pixels. Watershed seeds are placed at local minima of the structural edge map with a minimum inter-seed distance of $d_{\min} = 2$ pixels, and micro-regions smaller than $s_{\min} = 10$ pixels are absorbed into their most similar DINO neighbor.

\textbf{Evaluation Metrics.} We evaluate Part-DINO with three standard metrics. Normalized Mutual Information (NMI) measures the mutual information between predicted cluster assignments and ground-truth labels. A high NMI indicates that predicted clusters are strongly aligned with ground-truth parts. Adjusted Rand Index (ARI) measures whether pixel pairs from the same ground-truth part are assigned the same cluster, and pairs from different parts are assigned different clusters. ARI is adjusted for chance, so a random assignment scores near zero. Normalized Mean Error (NME) is calculated as the distance between centroid of each predicted region and measure its Euclidean distance to the nearest ground-truth keypoint (landmark such as eye or beak), and report the mean of these distances normalized by image size. Lower NME implies better centroid alignment.
\begin{figure*}[tp]
\centering
\setlength{\tabcolsep}{2pt} 
\renewcommand{\arraystretch}{0.5} 
\resizebox{\textwidth}{!}{

\input{tables/main_vis}
}
\caption{Qualitative comparison across CUB ($K{=}16$), CelebA ($K{=}16$), PartImageNet-S ($K{=}50$), and PartImageNet-O ($K{=}50$). MPAE is an unsupervised method that requires dataset-specific training, while \cite{amir2021deep}, \cite{amir2021deep} + SAM, and our Part-DINO are all training-free.}
\label{fig:qualitative_comparison}
\end{figure*}
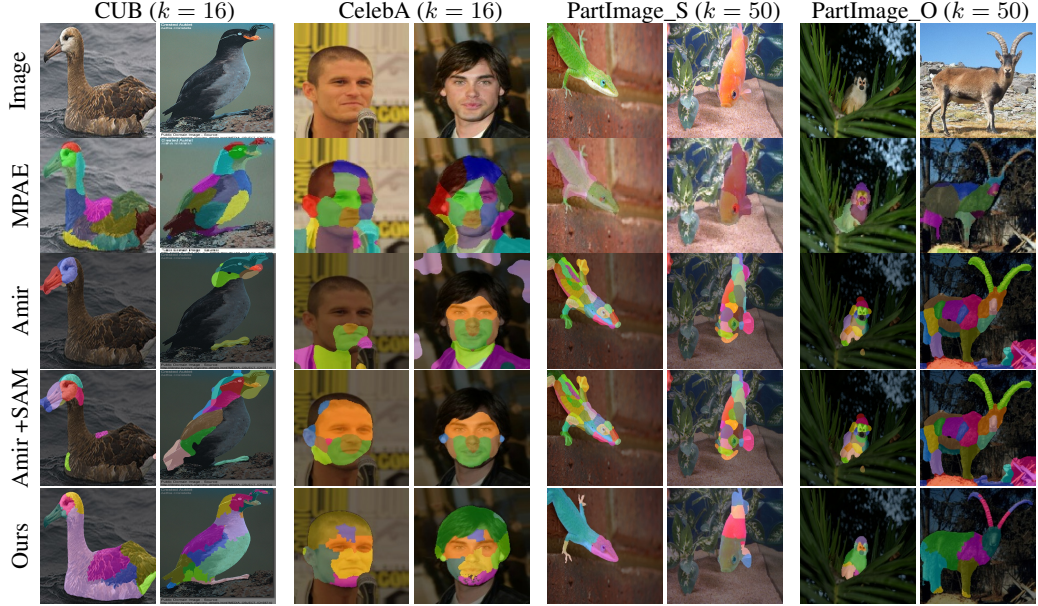

\textbf{Results.}
Tables \ref{tab:cub_benchmark} and \ref{tab:celeba_benchmark} show our quantitative results on the fine-grained CUB and CelebA datasets across  part granularities ($K \in \{4, 8, 16\}$). 
On CUB, Part-DINO outperforms the patch-based training-free baseline across all $K$, improving NMI (56.92$\rightarrow$65.45) and ARI (35.10$\rightarrow$46.00) at K=16. Despite requiring no training, Part-DINO matches or exceeds several weakly-supervised and unsupervised methods \citep{huang2020interpretable, pdisconet, xia2024unsupervised} at $K=16$. On CelebA, where the patch-based baseline collapses (ARI $<$ 0.1 across all $K$), Part-DINO outperforms the unsupervised GANSeg at $k=4$ (ARI: 32.20 vs. 28.06), though trained methods remain stronger.
Table~\ref{tab:partimagenet_horizontal} shows results on the PartImageNet OOD (PartImageNet$_O$) and Segmentation (PartImageNet$_S$) splits. On PartImageNet$_O$, Part-DINO achieves consistently higher ARI but slightly lower NMI compared to patch-based baselines. This reflects a known trade-off: NMI favors over-segmentation into many small, pure clusters, while ARI penalizes such fragmentation. By grouping spatially coherent regions, Part-DINO prioritizes structurally consistent parts, leading to higher ARI at the cost of marginally lower NMI. On PartImageNet$_S$, Part-DINO improves over the training-free baseline on both NMI and ARI for  K $\in \{8,25,50\}$, maintains an ARI above 66, and outperforms the fully trained SOTA MPAE at K = 8 on both metrics (37.21/66.10 vs. 33.51/65.05) despite no training.


\section{Analysis}
\label{sec:ablation}
\textbf{Unit (Patches vs Regions):} 
To isolate the impact of the atomic unit, we test all combinations of patches and regions for prototype construction and label assignment.
As shown in Table~\ref{tab:unit_ablation_cub_k4}, improving only the prototype quality while keeping patches as the assignment unit (row 2) yields marginal gains over the baseline (+6.50 NMI, +2.14 ARI). This is expected: while the target prototypes are cleaner, the patch-based assignment still blurs features across part boundaries. Improving only the assignment unit while keeping patch-based prototypes (row 3) yields a larger gain (+9.97 NMI, +4.17 ARI), as geometric region produces clean query features even when the targets are noisy. Combining both changes (row 4), we match region features with region prototypes, reducing mixing on both sides of the assignment and improving NMI (32.83 $\rightarrow$ 53.93) and ARI (13.96 $\rightarrow$ 23.01).

\begin{table}[tp]
    \centering
    
    \begin{minipage}[t]{0.5\textwidth}
        \centering
        \caption{Impact of the atomic unit on CUB ($k=4$). Combining region-based assignment with region-based prototypes (row 4) yields the largest improvement across all metrics.}
        \label{tab:unit_ablation_cub_k4}
        \renewcommand{\arraystretch}{1.2} 
        \resizebox{\linewidth}{!}{ 
            \begin{tabular}{llccc}
\toprule
Unit & Prototypes & NME ($\downarrow$) & NMI ($\uparrow$) & ARI ($\uparrow$) \\
\midrule
\rowcolor{lightgray!45} Patches & Patch-based & 9.22 & 32.83 & 13.96 \\
Patches & Region-based & 8.30 & 39.33 & 16.10 \\
\rowcolor{lightgray!45} Regions & Patch-based &9.02 & 42.80& 18.13 \\
Regions & Region-based & 8.21 & 53.93 & 23.01 \\
\bottomrule
\end{tabular}
        }
    \end{minipage}\hfill
    \begin{minipage}[t]{0.45\textwidth}
        \centering
        \caption{Region generation methods on CUB ($k=16$). Even at coarser granularity, RAMP + Watershed outperforms fine-grained SLIC.}
        \label{tab:superpixel_methods_regions}
        \renewcommand{\arraystretch}{1.2}
        \resizebox{0.9\linewidth}{!}{ 
            \begin{tabular}{lcccc}
            \toprule
            Method & Regions & NME $\downarrow$ & NMI $\uparrow$ & ARI $\uparrow$ \\
            \midrule
            Patches & \cellcolor{lightgray!45}-- & \cellcolor{lightgray!45}4.85 & \cellcolor{lightgray!45}49.03 & \cellcolor{lightgray!45}22.54 \\
            \midrule
            \multirow{2}{*}{SLIC} 
            & 250 & 5.00 & 58.55 & 36.94 \\
            & \cellcolor{lightgray!45}500 & \cellcolor{lightgray!45}4.99 & \cellcolor{lightgray!45}60.78 & \cellcolor{lightgray!45}38.64 \\
            \midrule
            \multirow{2}{*}{\shortstack{RAMP \\ + Watershed}}
            & 142 & 4.44 & 61.96 & 43.13 \\
            & \cellcolor{lightgray!45}500 & \cellcolor{lightgray!45}4.35 & \cellcolor{lightgray!45}65.45 & \cellcolor{lightgray!45}46.00 \\
            \bottomrule
            \end{tabular}
        }
    \end{minipage}
    
\end{table}

\begin{table}
\small
\captionof{table}{ Quantitative comparison on PartImageNet Out-of-Distribution (PartImageNet$_{O}$) and Segmentation (PartImageNet$_{S}$) splits across $K \in \{8, 25, 50\}$. Part-DINO shows consistent improvements over the training-free baseline on PartImageNet$_{S}$ across all granularities and metrics, and remains competitive on PartImageNet$_{O}$. $\dagger$ indicates replicated with SAM foreground mask}
\label{tab:partimagenet_horizontal}
\centering
\resizebox{\linewidth}{!}{
\input{tables/main_partimagenet}
}
\end{table}

\textbf{Region Generation:}  We compare RAMP-based Watershed \cite{akbas_ahuja,akbas2014low} against SLIC \cite{slic}, keeping all other components fixed. As shown in Table~\ref{tab:superpixel_methods_regions}, both methods improve over the patch baseline. At the same granularity of 500 regions, RAMP + Watershed improves ARI by +7.36 over SLIC, and even at a coarser granularity (142 regions) outperforms fine-grained SLIC (500 regions). The difference is visible in Figure~\ref{fig:ablation_watershed}: SLIC impose an explicit compactness constraint that trades boundary adherence for spatial uniformity. RAMP + Watershed places boundaries along intensity gradients, resulting in regions that respect object geometry.

\textbf{Region Granularity.} To study the effects of region granularity, we vary the minimum inter-seed distance $d_{\min}$. As shown in Figure~\ref{fig:num_regions_performance}, finer regions improve part discovery. Decreasing $d_{\min}$ from 15 to 2 increases the average number of regions from 22 to 500, improving all metrics (e.g ARI 25.21 $\rightarrow$ 46.00). The performance drop with coarser regions occurs because large regions span multiple parts, pooling features across part contours. In contrast, finer regions ensure that each unit remains contained within a single part, thus preserving the boundary precision.

\textbf{Structural vs. Semantic Edge Map.} Our region generation relies on structural edge map ($E_{\text{edge}}$) derived from the RAMP transform, explicitly avoiding the use of DINO semantic boundaries. As shown in Table~\ref{tab:ablation_edges}, compared to a baseline semantic similarity between patch features, structural edges consistently perform better across all granularities. The performance the gap widening as $K$ increases, at $K=16$, structural edges improve NMI by +9.06 and ARI by +7.05. Figure~\ref{fig:ablation_edges} shows semantic edges are blurred at object boundaries, producing regions that bleed across physical contours.

\textbf{Computational Efficiency.} We compare the computational efficiency of our method with \citep{amir2021deep}. Both share an identical DINOv2 backbone (\textasciitilde40\,ms) and SAM3 masking (\textasciitilde560\,ms). The main difference lies in Part-DINO's region-construction pipeline: RAMP (32\,ms), watershed (25\,ms), merging (6\,ms), and pooling (1\,ms), which introduces a negligible total overhead of 72\,ms per image. For this minimal cost, our method results in +8.6 NMI and +10.9 ARI at $d_{\min}=2$. More detailed comparison of each component across different $d_{min}$ is shown in Appendix Sec.\ref{supp_sec:computational_cost} and Tab.\ref{tab:timing_and_metrics}.

\begin{figure}[tp]
    \centering
    \begin{minipage}[c]{0.5\textwidth}
        \centering
        \captionof{table}{Structural vs. Semantic edge maps on CUB. Regions derived from purely geometric edges consistently outperform those derived from DINOv2 semantic edges across all $K$, with the gap widening at finer granularities (e.g., +9.06 NMI and +7.05 ARI at $K{=}16$).}
        \label{tab:ablation_edges}
        \renewcommand{\arraystretch}{1.2}
        \resizebox{0.9\linewidth}{!}{ 
            \begin{tabular}{clccc}
            \toprule
            $K$ & Edges & NME ($\downarrow$) & NMI ($\uparrow$) & ARI ($\uparrow$) \\
            \midrule 
            \multirow{2}{*}{4} & \cellcolor{lightgray!45}Semantic & \cellcolor{lightgray!45}8.23 & \cellcolor{lightgray!45}48.73 & \cellcolor{lightgray!45}19.84 \\
              & Geometry & 8.21 & 53.93 & 23.01 \\
            \midrule
            \multirow{2}{*}{8} & \cellcolor{lightgray!45}Semantic & \cellcolor{lightgray!45}6.11 & \cellcolor{lightgray!45}50.91 & \cellcolor{lightgray!45}24.60 \\
              & Geometry & 6.00 & 61.84 & 37.92 \\
            \midrule
            \multirow{2}{*}{16} & \cellcolor{lightgray!45}Semantic & \cellcolor{lightgray!45}4.63 & \cellcolor{lightgray!45}56.39 & \cellcolor{lightgray!45}38.95 \\
               & Geometry & 4.35 & 65.45 & 46.00 \\
            \bottomrule
            \end{tabular}
        }
    \end{minipage}\hfill
    \begin{minipage}[c]{0.46\textwidth}
        \centering
        \includegraphics[width=\linewidth]{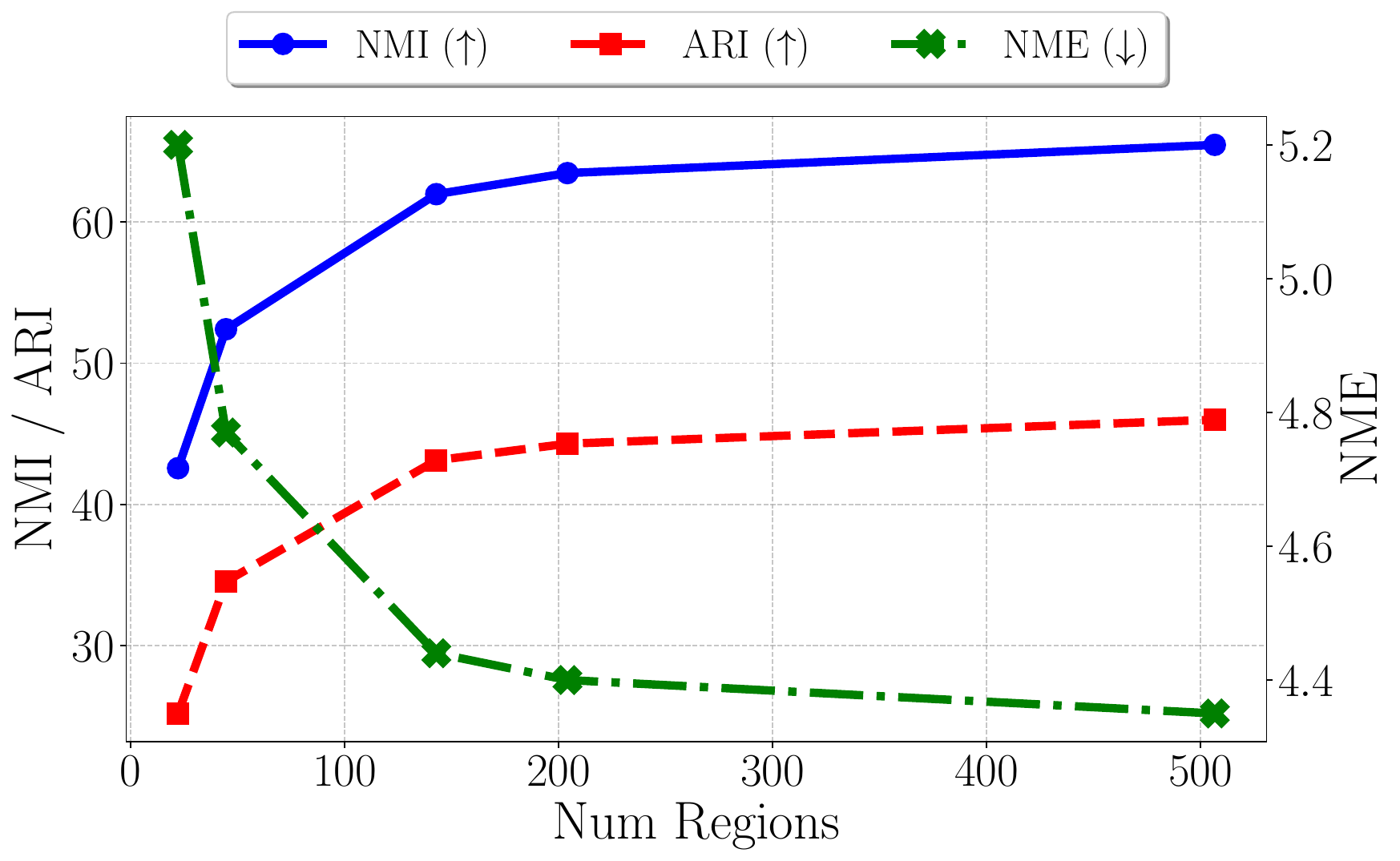}
        \caption{Effect of region granularity on CUB ($K{=}16$). Finer granularities improve part discovery across all metrics, with performance saturating around 300 regions.}
        \label{fig:num_regions_performance}
    \end{minipage}
\end{figure}

\begin{figure*}[tp]
\centering

\begin{minipage}[t]{0.54\textwidth}
\centering
\setlength{\tabcolsep}{2pt}
\renewcommand{\arraystretch}{0.5}
\resizebox{\linewidth}{!}{
\begin{tabular}{ccccc}
    & \small Original & \small Edge Map & \small Regions & \small Overlay \\
    
    \raisebox{-.5\height}{\rotatebox{90}{\small \textit{Semantic}} }&
    \raisebox{-.5\height}{\includegraphics[width=1.8cm, height=1.8cm, keepaspectratio]{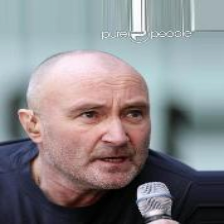}} &
    \raisebox{-.5\height}{\includegraphics[width=1.8cm, height=1.8cm, keepaspectratio]{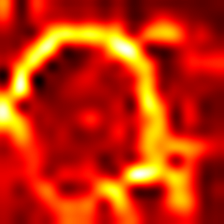}} &
    \raisebox{-.5\height}{\includegraphics[width=1.8cm, height=1.8cm, keepaspectratio]{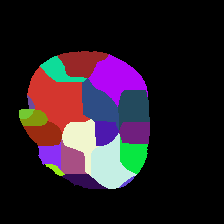}} &
    \raisebox{-.5\height}{\includegraphics[width=1.8cm, height=1.8cm, keepaspectratio]{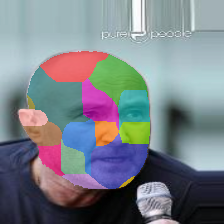}} \\
    
    \raisebox{-.5\height}{\rotatebox{90}{\small \textit{Structural}}} &
    \raisebox{-.5\height}{\includegraphics[width=1.8cm, height=1.8cm, keepaspectratio]{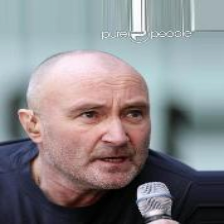}} &
    \raisebox{-.5\height}{\includegraphics[width=1.8cm, height=1.8cm, keepaspectratio]{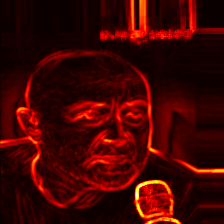}} &
    \raisebox{-.5\height}{\includegraphics[width=1.8cm, height=1.8cm, keepaspectratio]{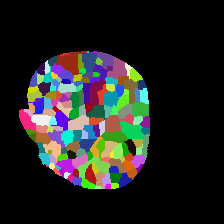}} &
    \raisebox{-.5\height}{\includegraphics[width=1.8cm, height=1.8cm, keepaspectratio]{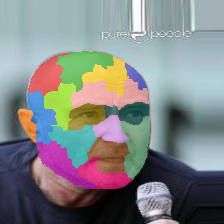}} \\
\end{tabular}
}
\subcaption{Structural vs. Semantic Edge Map.}
\label{fig:ablation_edges}
\end{minipage}%
\hfill
\begin{minipage}[t]{0.38\textwidth}
\centering
\setlength{\tabcolsep}{2pt}
\renewcommand{\arraystretch}{0.5}
\resizebox{\linewidth}{!}{
\begin{tabular}{cccc}
    &  Original &  Regions &  Overlay \\
    
    \raisebox{-.5\height}{\rotatebox{90}{ \textit{SLIC}}} &
    \raisebox{-.5\height}{\includegraphics[width=3.2cm, height=3.2cm, keepaspectratio]{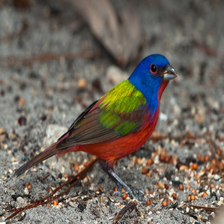}} &
    \raisebox{-.5\height}{\includegraphics[width=3.2cm, height=3.2cm, keepaspectratio]{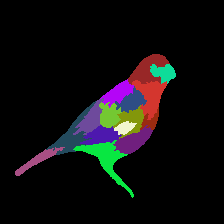}} &
    \raisebox{-.5\height}{\includegraphics[width=3.2cm, height=3.2cm, keepaspectratio]{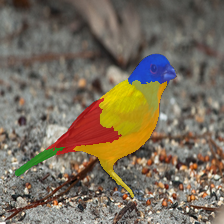}} \\
    
    \raisebox{-.5\height}{\rotatebox{90}{ \textit{RAMP + Watershed}}} &
    \raisebox{-.5\height}{\includegraphics[width=3.2cm, height=3.2cm, keepaspectratio]{images/slic/00_original.png}} &
    \raisebox{-.5\height}{\includegraphics[width=3.2cm, height=3.2cm, keepaspectratio]{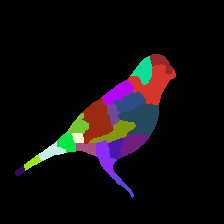}} &
    \raisebox{-.5\height}{\includegraphics[width=3.2cm, height=3.2cm, keepaspectratio]{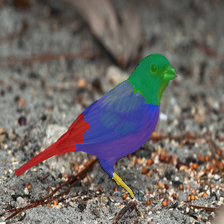}} \\
\end{tabular}
}
\subcaption{Watershed vs. SLIC.}
\label{fig:ablation_watershed}
\end{minipage}
\caption{Qualitative experiments on region generation. (a) The structural edges from RAMP align with object contours, while DINO's semantic edges are blurred. (b) Watershed regions are separable (head, body, wing, and tail), while SLIC prioritizes spatial compactness and cuts across them.}
\label{fig:ablation_combined}
\end{figure*}

\section{Discussion}
We have presented Part-DINO, a training-free framework motivated by an empirical finding: pixel-space mixing at boundary patches propagates into feature space. Across 26k foreground patches, feature entropy rises and the top-2 assignment margin falls as patch–boundary overlap increases, making the ViT patch the bottleneck for spatially precise part discovery. Part-DINO addresses this by replacing the patch with regions whose boundaries come from low-level image segmentation alone, independent of DINO. We pool DINO features within each region and use them only to group regions into parts, not to place boundaries. Part-DINO outperforms the patch-based training-free baseline on NME and ARI across all four benchmarks and granularities, and matches or exceeds several training-based methods.

We present \textbf{two limitations}. (1) Fixed part granularity. Part-DINO requires the number of parts ($K$) to be specified in advance. While this is a common assumption across all part discovery methods, it limits applicability to settings where the correct granularity is unknown or varies across object categories. (2) Dependence on contrast. Because our region boundaries are derived from the RAMP transform, which relies on local intensity contrast, the edge response weakens in low-contrast scenes, uniform lighting or minimal texture variation between parts, leading to under-segmentation. Conversely, strong illumination gradients or cast shadows can introduce spurious edges that do not correspond to part boundaries.

\textbf{Future work.} Part-DINO discovers parts as unlabeled clusters. A natural extension is to assign semantic labels to these discovered regions by leveraging vision-language models such as CLIP, bridging part discovery and open-vocabulary part segmentation. 
\textbf{Broader Impact.} Positively, our training-free approach reduces the computational energy footprint typically required for dataset-specific part discovery. Negatively, enhanced unsupervised localization capabilities could be intentionally misused for unauthorized biometric surveillance.

\section*{Acknowledgments}

This work was supported in part by the Office of Naval Research under Grant N00014-24-1-2169, the IBM-Illinois Discovery Accelerator Institute (IIDAI), and the Amazon--Illinois Center on AI for Interactive Conversational Experiences. This research also utilized resources from the NCSA Delta system, supported by the National Science Foundation under Award OAC-2005572.
{
    \small
    \bibliographystyle{ieeenat_fullname}
    \bibliography{main}
}

\newpage
\input{checklist.tex} 
\newpage

\appendix
\section*{Appendix}
In this Appendix, we provide an extended analysis of the computational efficiency of our framework (Appendix \ref{supp_sec:computational_cost}), further empirical validation of the geometric mixing that corrupts ViT boundary patches (Appendix \ref{supp_sec:Boundary}), and extended qualitative comparisons evaluating structural versus semantic boundaries (Appendix \ref{supp_sec:Semantic vs. Structural Boundaries}) as well as Watershed versus SLIC region generation across datasets (Appendix \ref{supp:Extended Qualitative Analysis on Region Generation: SLIC vs. Watershed}). Lastly, we demonstrate that region generation remains effective even without the DINO-based micro-region merge (Appendix \ref{sec:supp_merge}).

\section{Computational Efficiency: Extended Analysis}
\label{supp_sec:computational_cost}

\paragraph{Setup.}
We report means over the CUB test split at resolution $448{\times}448$ on a single NVIDIA A40 GPU with 48 CPU cores (no batching). The \citep{amir2021deep} patch baseline shares the identical backbone and SAM3 segmenter call, only the feature-extraction and assignment stages differ.

\paragraph{Stage-level breakdown.}
Table~\ref{tab:timing_and_metrics} decomposes latency into seven stages. The \emph{shared backbone} (DINO forward + SAM3 mask) accounts for approximately 600\,ms in both methods and represents the dominant cost. Among Part-DINO's method-specific stages, RAMP edge estimation and watershed seeding total \textasciitilde57\,ms and are independent of $d_{\min}$. The merge stage scales mildly with the number of atomic regions: at $d_{\min}=2$ (538 regions, mean), merge takes 7.7\,ms; at $d_{\min}=15$ (23 regions), it takes 5.2\,ms. Region pooling via \texttt{torch.scatter\_add} requires \textasciitilde1\,ms regardless of granularity.

\section{ Boundary Analysis Details}
\label{supp_sec:Boundary}
In the main text, we hypothesize that the fixed-size square grid ($14 \times 14$ pixels) used by Vision Transformers acts as a fundamental bottleneck for spatial precision. Figure \ref{supp_fig:analysis} visualizes the empirical validation of this hypothesis through an analysis of $N = 26,459$ foreground patches from the PartImageNet dataset, examining how semantic representation degrades as a function of distance to physical object contours.

To rigorously evaluate this geometric relationship, we first constructed ground-truth part centroids using high-purity patches ($\pi > 0.95$) from the training set, anchoring them to isolated, unambiguous semantic parts. For each patch $p$, we calculated its Euclidean distance $d$ to the nearest ground-truth part boundary. We then mapped this distance against three key patch-level metrics:
\begin{enumerate}
    \item Spatial Purity ($\pi$): The fraction of the patch's foreground pixels belonging to its majority part label.
    \item Feature Entropy ($H$): The Shannon entropy of the patch's feature distances relative to the part-anchored centroids, measuring representational ambiguity.
    \item Top-2 Margin ($m$): The difference in similarity between the nearest and second-nearest part centroids, serving as a measure of assignment confidence.
\end{enumerate}

As illustrated by the distance curves in Figure \ref{supp_fig:analysis}, a patch's proximity to a boundary heavily dictates its semantic reliability. We observe a strong positive correlation ($\rho(d,\pi) = +0.71$) between boundary distance and spatial purity; patches directly on or adjacent to a contour unavoidably pool pixels from distinct parts. 

Crucially, this geometric mixing corrupts the underlying feature space. As patches move closer to the boundary (decreasing $d$), feature entropy rises ($\rho(d,H) = -0.38$) and the assignment margin steadily narrows ($\rho(d,m) = +0.36$). While the dataset-wide mean patch purity is high ($0.952$), the sharp degradation of these metrics at $d \approx 0$ quantitatively demonstrates that square ViT patches become inherently ambiguous exactly where precise boundary localization is required.

\section{Extended Qualitative Analysis: Semantic vs. Structural Boundaries}
\label{supp_sec:Semantic vs. Structural Boundaries}
In the main text, we established that extracting boundaries via semantic feature similarity leads to spatial imprecision, whereas relying on structural image contrast preserves object contours. Figures \ref{fig:supp_ablation_celeba}, \ref{fig:supp_ablation_cub}, and \ref{fig:supp_ablation_partimagenet} provide an extended qualitative comparison of these edge map modalities across various datasets.

As demonstrated in the figures, edges derived from DINO similarity inherently suffer from feature degradation near part transitions, causing the predicted regions to bleed across actual object boundaries. By contrast, our structural boundary formulation leverages the RAMP transform to isolate edge discontinuities independent of the DINO feature space. Across CelebA (Figure \ref{fig:supp_ablation_celeba}), and other complex object categories (Figures \ref{fig:supp_ablation_cub} and \ref{fig:supp_ablation_partimagenet}), the structural boundaries consistently reduce semantic bleeding artifacts and produce sharp parts.

\section{Extended Qualitative Analysis on Region Generation: SLIC vs. Watershed}
\label{supp:Extended Qualitative Analysis on Region Generation: SLIC vs. Watershed}
To construct atomic regions that do not arbitrarily cross object parts, the choice of the grouping algorithm is critical.Figures \ref{fig:supp_ablation_watershed_slic_1} and \ref{fig:supp_ablation_watershed_slic_2} provide an extended ablation comparing our chosen Watershed transform against SLIC superpixels across a variety of complex categories.

While SLIC clusters pixels based on a joint color-spatial space, it imposes a strict spatial compactness penalty. As seen in the SLIC rows of Figures \ref{fig:supp_ablation_watershed_slic_1} and \ref{fig:supp_ablation_watershed_slic_2}, when structural contours are weak or highly irregular, this compactness constraint dominates, forcing arbitrary, grid-like boundaries that cut directly across anatomical shapes. Conversely, our Watershed approach operates on the topographic valleys of the structural gradient. As shown in the Watershed rows, this allows the generated atomic regions to break free from spatial rigidity and seamlessly wrap around the true, natural contours of the physical objects.

\section{DINO-Free Region Generation}
\label{sec:supp_merge}

DINO enters region generation only through the micro-region merge (Sec.~3.1), which absorbs
each watershed region smaller than $s_{\min}$ pixels into its most similar neighbor. Merging
only removes boundaries and never places one. DINO therefore decides which watershed regions
are grouped, the same role it plays in region assignment (Sec.~3.3), while boundary placement
depends on $E_{\text{edge}}$ alone. To confirm that the gains do not depend on this step, we
disable it ($s_{\min}=0$), making region generation fully independent of DINO.

Table~\ref{tab:merge} reports the result on CUB at $K=16$. Disabling the merge adds 18.5
regions per image and changes NMI and NME by only 0.03, while ARI drops by 1.91. Even without
the merge, Part-DINO outperforms the patch-based baseline by +8.09 NMI and +8.82 ARI, with
0.51 lower NME. The gains therefore come from the DINO-independent boundaries, not from the
merge.

\begin{table}[h]
\centering\small
\caption{Effect of the DINO-based micro-region merge on CUB ($K=16$). With $s_{\min}=0$,
region generation uses no DINO features. Both Part-DINO rows come from the same run.
$^\dagger$ replicated with SAM foreground mask.}
\label{tab:merge}
\begin{tabular}{lcccc}
\toprule
Method & Regions/img & NME $\downarrow$ & NMI $\uparrow$ & ARI $\uparrow$ \\
\midrule
\citet{amir2021deep}$^\dagger$ & -- & 4.87 & 56.92 & 35.10 \\
Part-DINO ($s_{\min}=10$, default, rerun) & 506.7 & 4.33 & 64.98 & 45.83 \\
Part-DINO ($s_{\min}=0$, no merge) & 525.2 & 4.36 & 65.01 & 43.92 \\
\bottomrule
\end{tabular}
\end{table}

\begin{table}[ht]
\centering
\caption{Timing breakdown (ms), average regions, and clustering performance metrics.}
\label{tab:timing_and_metrics}
\resizebox{\textwidth}{!}{%
\begin{tabular}{@{}lrrrrrrrrrrrr@{}}
\toprule
& \multicolumn{8}{c}{Runtime Breakdown (ms)} & & \multicolumn{3}{c}{Metrics} \\
\cmidrule(lr){2-9} \cmidrule(l){11-13}
Method & DINO & SAM & RAMP & Wshed & Merge & Pool & Assign & Total & Avg. Reg. & NME ($\downarrow$) & NMI ($\uparrow$) & ARI ($\uparrow$) \\
\midrule
\rowcolor{lightgray!45}\citep{amir2021deep} (patch) & 40.0 & 552.4 & 0.0 & 0.0 & 0.0 & 6.0 & 1.1 & $599.5 $ & 1024 & 4.87 & 56.92 & 35.10 \\
Part-DINO ($d_{min}=15$) & 40.2 & 558.3 & 32.4 & 23.8 & 5.2 & 1.3 & 0.4 & $664.9 $ & 23.1  & 5.20 & 42.57 & 25.21 \\
\rowcolor{lightgray!45}Part-DINO ($d_{min}=5$) & 39.7 & 562.9 & 32.8 & 24.5 & 6.1 & 1.3 & 1.4 & $672.2 $ & 151.1 & 4.44 & 61.96 & 43.13 \\
Part-DINO ($d_{min}=2$) & 39.6 & 563.1 & 31.9 & 29.0 & 7.7 & 1.4 & 3.7 & $680.2 $ & 562.1 & 4.35 & 65.45 & 46.00 \\
\bottomrule
\end{tabular}%
}
\end{table}

\begin{figure*}
\centering
\setlength{\tabcolsep}{2pt}
\renewcommand{\arraystretch}{0.5}

\begin{tabular}{ccccc}
    & \small Original Image & \small Edge Map & \small Atomic Regions & \small Final Overlay \\

    \rotatebox{90}{\small \textit{Semantic}} &
    \raisebox{-.5\height}{\includegraphics[width=0.23\linewidth]{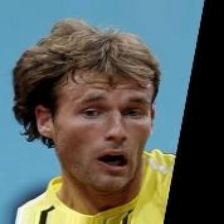}} &
    \raisebox{-.5\height}{\includegraphics[width=0.23\linewidth]{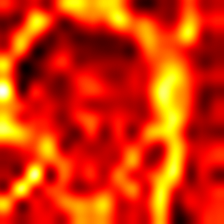}} &
    \raisebox{-.5\height}{\includegraphics[width=0.23\linewidth]{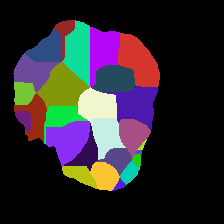}} &
    \raisebox{-.5\height}{\includegraphics[width=0.23\linewidth]{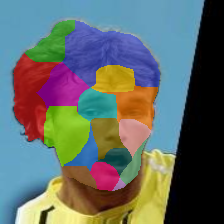}} \\
    \rotatebox{90}{\small \textit{Structural}} &
    \raisebox{-.5\height}{\includegraphics[width=0.23\linewidth]{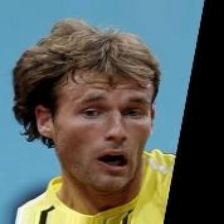}} & 
    \raisebox{-.5\height}{\includegraphics[width=0.23\linewidth]{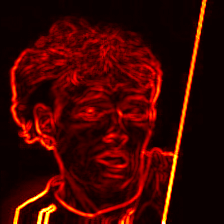}} &
    \raisebox{-.5\height}{\includegraphics[width=0.23\linewidth]{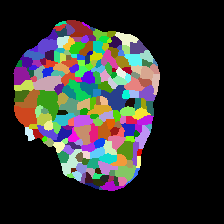}} &
    \raisebox{-.5\height}{\includegraphics[width=0.23\linewidth]{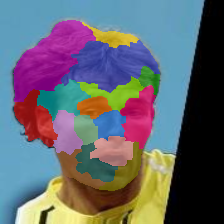}} \\
\end{tabular}

\caption{\textbf{Extended Qualitative Analysis (CelebA).} Comparison of edge map modalities. Structural boundaries consistently eliminate the semantic bleeding artifacts present in the patch-based baseline.}
\label{fig:supp_ablation_celeba}
\end{figure*}

\begin{figure*}
    \centering
    \includegraphics[width=0.32\linewidth]{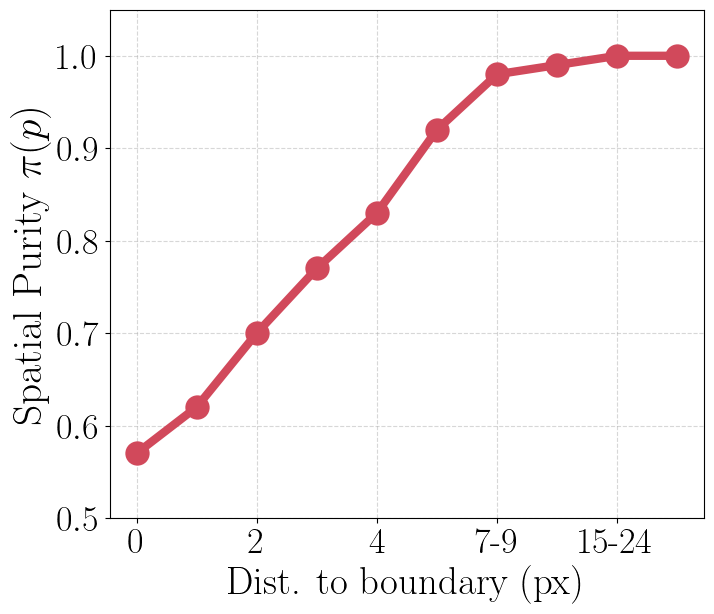}\hfill
    \includegraphics[width=0.32\linewidth]{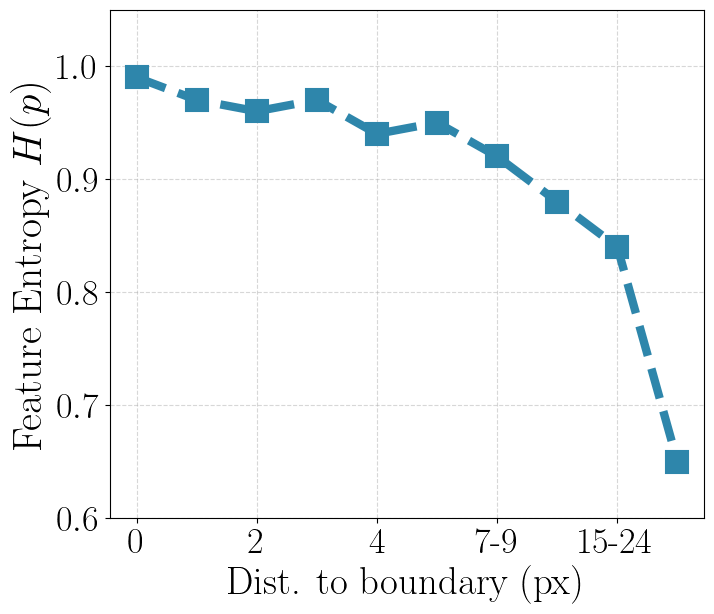}\hfill
    \includegraphics[width=0.32\linewidth]{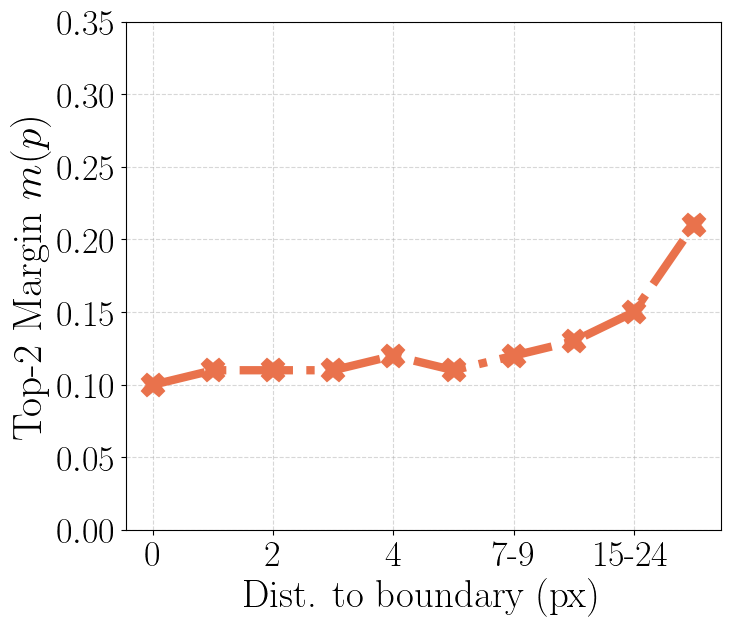}
    \caption{\textbf{Quantifying the boundary bottleneck.} Across 26,459 foreground patches, all three reliability metrics correlate monotonically with distance to the nearest part boundary: spatial purity rises $\rho(d, \pi) = +0.71$), while feature entropy falls (($\rho(d, H) = -0.38$), and top-2 assignment margin grows ($\rho(d, m)$ = +0.36), confirming that geometric mixing propagates into feature space}
    \label{supp_fig:analysis}
\end{figure*}

\begin{figure*}[ht]
\centering
\setlength{\tabcolsep}{2pt}
\renewcommand{\arraystretch}{0.5}

\begin{tabular}{ccccc}
    & \small Original Image & \small Edge Map & \small Atomic Regions & \small Final Overlay \\

    \midrule
    \rotatebox{90}{\small \textit{Semantic}} &
    \raisebox{-.5\height}{\includegraphics[width=0.23\linewidth]{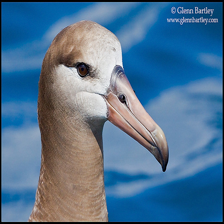}} &
    \raisebox{-.5\height}{\includegraphics[width=0.23\linewidth]{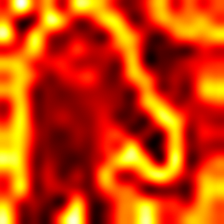}} &
    \raisebox{-.5\height}{\includegraphics[width=0.23\linewidth]{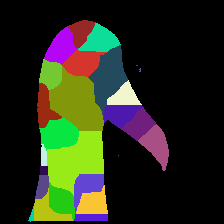}} &
    \raisebox{-.5\height}{\includegraphics[width=0.23\linewidth]{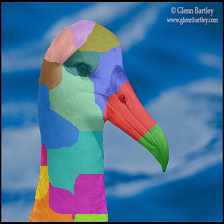}} \\
    \rotatebox{90}{\small \textit{Structural}} &
    \raisebox{-.5\height}{\includegraphics[width=0.23\linewidth]{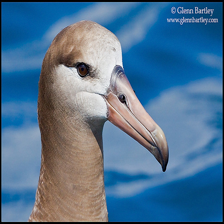}} & 
    \raisebox{-.5\height}{\includegraphics[width=0.23\linewidth]{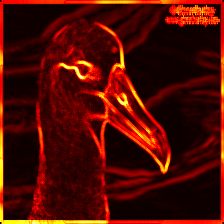}} &
    \raisebox{-.5\height}{\includegraphics[width=0.23\linewidth]{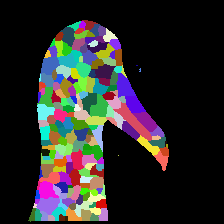}} &
    \raisebox{-.5\height}{\includegraphics[width=0.23\linewidth]{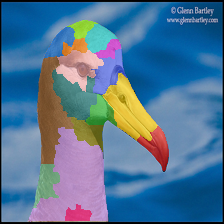}} \\

    \midrule
    \rotatebox{90}{\small \textit{Semantic}} &
    \raisebox{-.5\height}{\includegraphics[width=0.23\linewidth]{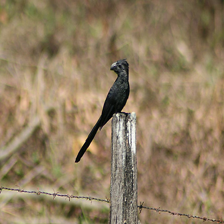}} &
    \raisebox{-.5\height}{\includegraphics[width=0.23\linewidth]{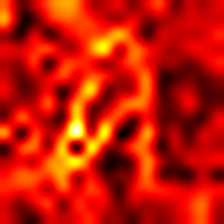}} &
    \raisebox{-.5\height}{\includegraphics[width=0.23\linewidth]{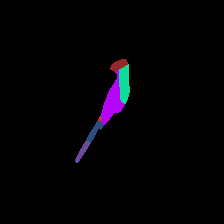}} &
    \raisebox{-.5\height}{\includegraphics[width=0.23\linewidth]{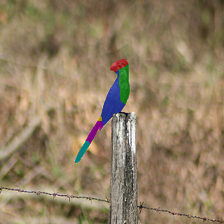}} \\
    \rotatebox{90}{\small \textit{Structural}} &
    \raisebox{-.5\height}{\includegraphics[width=0.23\linewidth]{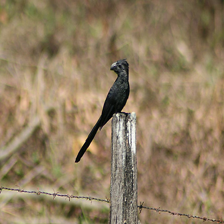}} & 
    \raisebox{-.5\height}{\includegraphics[width=0.23\linewidth]{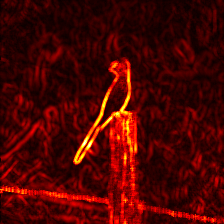}} &
    \raisebox{-.5\height}{\includegraphics[width=0.23\linewidth]{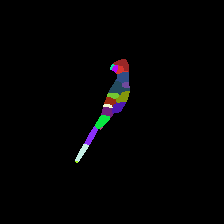}} &
    \raisebox{-.5\height}{\includegraphics[width=0.23\linewidth]{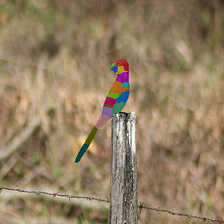}} \\
\end{tabular}

\caption{\textbf{Extended Qualitative Analysis (CUB).} Comparison of edge map modalities on fine-grained bird categories, highlighting precision improvements in part boundaries.}
\label{fig:supp_ablation_cub}
\end{figure*}

\begin{figure*}[ht]
\centering
\setlength{\tabcolsep}{2pt}
\renewcommand{\arraystretch}{0.5}

\begin{tabular}{ccccc}
    & \small Original Image & \small Edge Map & \small Atomic Regions & \small Final Overlay \\
    
    \midrule
    \rotatebox{90}{\small \textit{Semantic}} &
    \raisebox{-.5\height}{\includegraphics[width=0.23\linewidth]{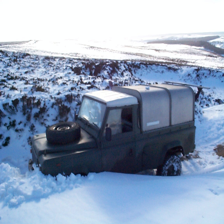}} &
    \raisebox{-.5\height}{\includegraphics[width=0.23\linewidth]{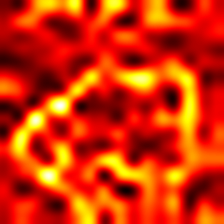}} &
    \raisebox{-.5\height}{\includegraphics[width=0.23\linewidth]{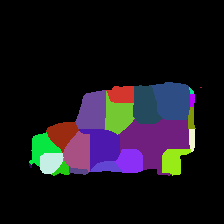}} &
    \raisebox{-.5\height}{\includegraphics[width=0.23\linewidth]{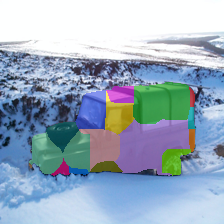}} \\
    \rotatebox{90}{\small \textit{Structural}} &
    \raisebox{-.5\height}{\includegraphics[width=0.23\linewidth]{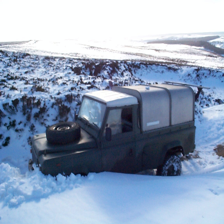}} & 
    \raisebox{-.5\height}{\includegraphics[width=0.23\linewidth]{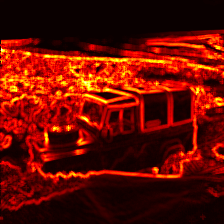}} &
    \raisebox{-.5\height}{\includegraphics[width=0.23\linewidth]{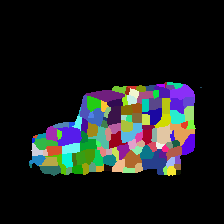}} &
    \raisebox{-.5\height}{\includegraphics[width=0.23\linewidth]{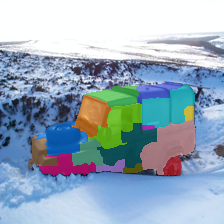}} \\

    \midrule
    \rotatebox{90}{\small \textit{Semantic}} &
    \raisebox{-.5\height}{\includegraphics[width=0.23\linewidth]{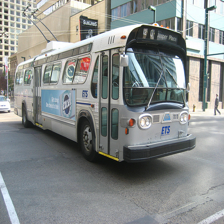}} &
    \raisebox{-.5\height}{\includegraphics[width=0.23\linewidth]{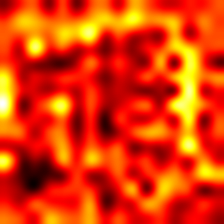}} &
    \raisebox{-.5\height}{\includegraphics[width=0.23\linewidth]{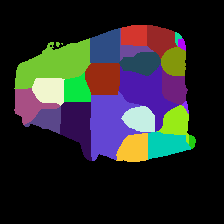}} &
    \raisebox{-.5\height}{\includegraphics[width=0.23\linewidth]{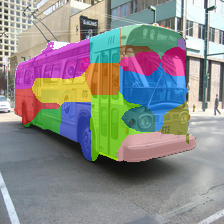}} \\
    \rotatebox{90}{\small \textit{Structural}} &
    \raisebox{-.5\height}{\includegraphics[width=0.23\linewidth]{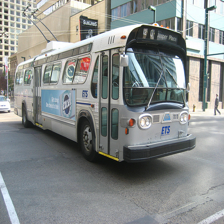}} & 
    \raisebox{-.5\height}{\includegraphics[width=0.23\linewidth]{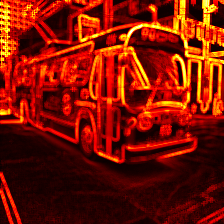}} &
    \raisebox{-.5\height}{\includegraphics[width=0.23\linewidth]{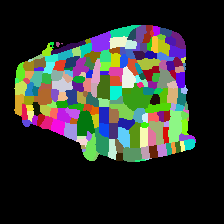}} &
    \raisebox{-.5\height}{\includegraphics[width=0.23\linewidth]{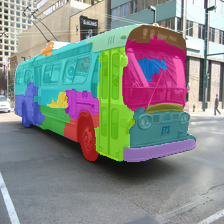}} \\

    \midrule
    \rotatebox{90}{\small \textit{Semantic}} &
    \raisebox{-.5\height}{\includegraphics[width=0.23\linewidth]{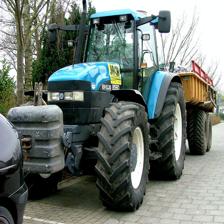}} &
    \raisebox{-.5\height}{\includegraphics[width=0.23\linewidth]{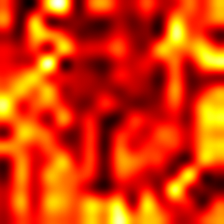}} &
    \raisebox{-.5\height}{\includegraphics[width=0.23\linewidth]{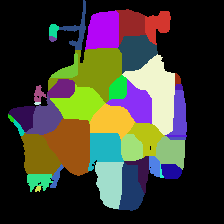}} &
    \raisebox{-.5\height}{\includegraphics[width=0.23\linewidth]{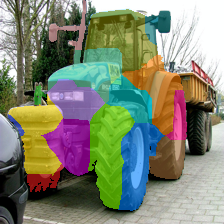}} \\
    \rotatebox{90}{\small \textit{Structural}} &
    \raisebox{-.5\height}{\includegraphics[width=0.23\linewidth]{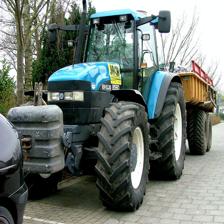}} & 
    \raisebox{-.5\height}{\includegraphics[width=0.23\linewidth]{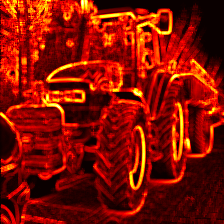}} &
    \raisebox{-.5\height}{\includegraphics[width=0.23\linewidth]{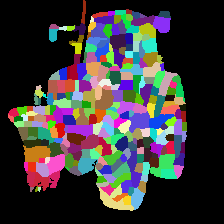}} &
    \raisebox{-.5\height}{\includegraphics[width=0.23\linewidth]{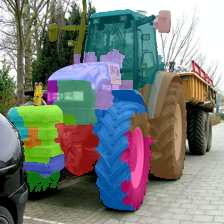}} \\
\end{tabular}

\caption{\textbf{Extended Qualitative Analysis (PartImageNet).} Comparison of structural vs. semantic edge modalities on Out-of-Distribution (OOD) and Segmentation (Seg) examples. Structural boundaries demonstrate robust geometric alignment regardless of semantic category.}
\label{fig:supp_ablation_partimagenet}
\end{figure*}

\begin{figure*}[ht]
\centering
\setlength{\tabcolsep}{4pt}
\renewcommand{\arraystretch}{0.5}

\begin{tabular}{cccc}
    & \small Original Image & \small Atomic Regions & \small Final Overlay \\
    
    \midrule
    \rotatebox{90}{\small \textit{SLIC}} &
    \raisebox{-.5\height}{\includegraphics[width=0.22\linewidth]{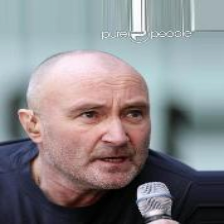}} &
    \raisebox{-.5\height}{\includegraphics[width=0.22\linewidth]{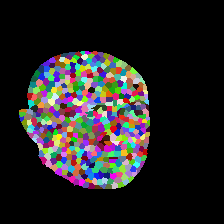}} &
    \raisebox{-.5\height}{\includegraphics[width=0.22\linewidth]{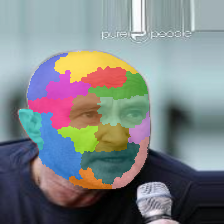}} \\
    \rotatebox{90}{\small \textit{Watershed}} &
    \raisebox{-.5\height}{\includegraphics[width=0.22\linewidth]{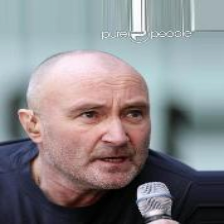}} & 
    \raisebox{-.5\height}{\includegraphics[width=0.22\linewidth]{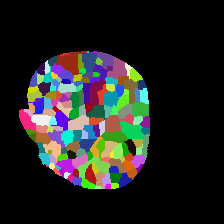}} &
    \raisebox{-.5\height}{\includegraphics[width=0.22\linewidth]{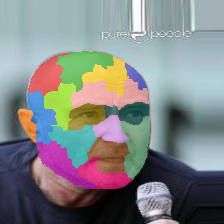}} \\

    \midrule
    \rotatebox{90}{\small \textit{SLIC}} &
    \raisebox{-.5\height}{\includegraphics[width=0.22\linewidth]{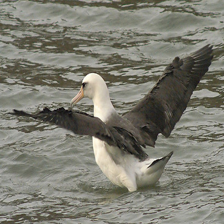}} &
    \raisebox{-.5\height}{\includegraphics[width=0.22\linewidth]{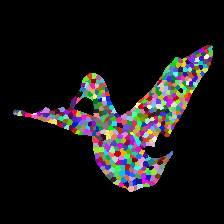}} &
    \raisebox{-.5\height}{\includegraphics[width=0.22\linewidth]{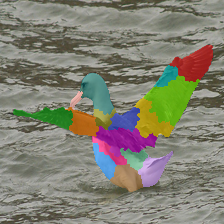}} \\
    \rotatebox{90}{\small \textit{Watershed}} &
    \raisebox{-.5\height}{\includegraphics[width=0.22\linewidth]{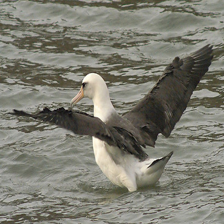}} & 
    \raisebox{-.5\height}{\includegraphics[width=0.22\linewidth]{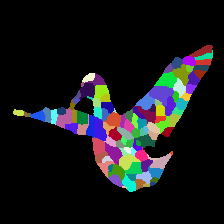}} &
    \raisebox{-.5\height}{\includegraphics[width=0.22\linewidth]{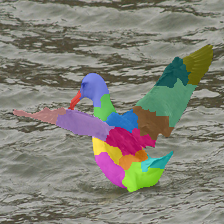}} \\
\end{tabular}

\caption{\textbf{Extended Ablation on Region Generation (Part 1).} SLIC enforces arbitrary spatial compactness, forcing grid-like boundaries that ignore true anatomical shapes. In contrast, our Watershed approach naturally follows the topographic valleys of the structural gradient, seamlessly wrapping around physical object contours.}
\label{fig:supp_ablation_watershed_slic_1}
\end{figure*}

\begin{figure*}[tp]
\centering
\setlength{\tabcolsep}{4pt}
\renewcommand{\arraystretch}{0.5}

\begin{tabular}{cccc}
    & \small Original Image & \small Atomic Regions & \small Final Overlay \\

    \midrule
    \rotatebox{90}{\small \textit{SLIC}} &
    \raisebox{-.5\height}{\includegraphics[width=0.22\linewidth]{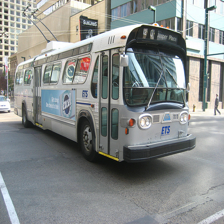}} &
    \raisebox{-.5\height}{\includegraphics[width=0.22\linewidth]{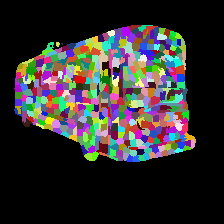}} &
    \raisebox{-.5\height}{\includegraphics[width=0.22\linewidth]{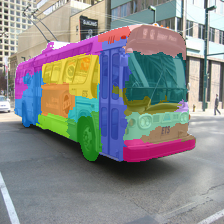}} \\
    \rotatebox{90}{\small \textit{Watershed}} &
    \raisebox{-.5\height}{\includegraphics[width=0.22\linewidth]{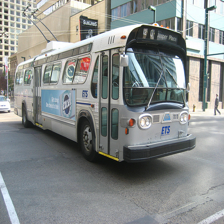}} & 
    \raisebox{-.5\height}{\includegraphics[width=0.22\linewidth]{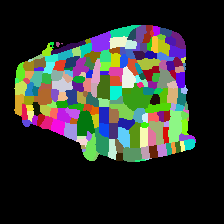}} &
    \raisebox{-.5\height}{\includegraphics[width=0.22\linewidth]{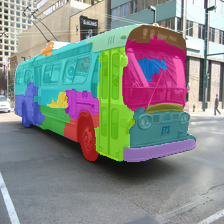}} \\

    \midrule
    \rotatebox{90}{\small \textit{SLIC}} &
    \raisebox{-.5\height}{\includegraphics[width=0.22\linewidth]{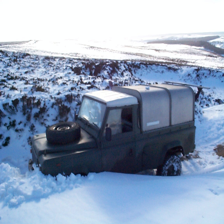}} &
    \raisebox{-.5\height}{\includegraphics[width=0.22\linewidth]{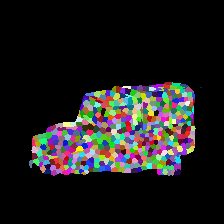}} &
    \raisebox{-.5\height}{\includegraphics[width=0.22\linewidth]{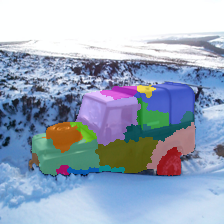}} \\
    \rotatebox{90}{\small \textit{Watershed}} &
    \raisebox{-.5\height}{\includegraphics[width=0.22\linewidth]{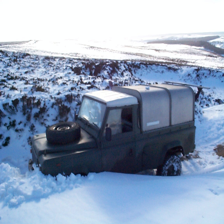}} & 
    \raisebox{-.5\height}{\includegraphics[width=0.22\linewidth]{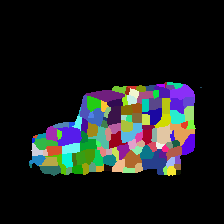}} &
    \raisebox{-.5\height}{\includegraphics[width=0.22\linewidth]{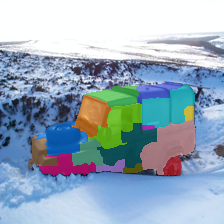}} \\

    \midrule
    \rotatebox{90}{\small \textit{SLIC}} &
    \raisebox{-.5\height}{\includegraphics[width=0.22\linewidth]{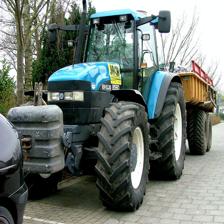}} &
    \raisebox{-.5\height}{\includegraphics[width=0.22\linewidth]{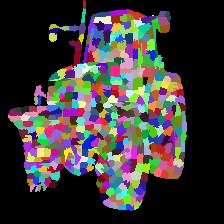}} &
    \raisebox{-.5\height}{\includegraphics[width=0.22\linewidth]{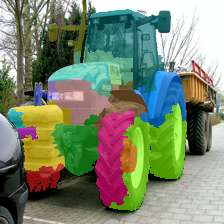}} \\
    \rotatebox{90}{\small \textit{Watershed}} &
    \raisebox{-.5\height}{\includegraphics[width=0.22\linewidth]{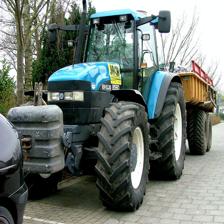}} & 
    \raisebox{-.5\height}{\includegraphics[width=0.22\linewidth]{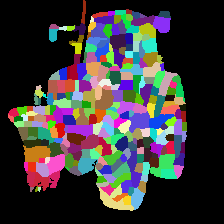}} &
    \raisebox{-.5\height}{\includegraphics[width=0.22\linewidth]{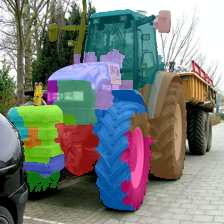}} \\
\end{tabular}
\caption{\textbf{Extended Ablation on Region Generation (Part 2).} Continued comparison across complex PartImageNet categories, demonstrating the robustness of Watershed boundaries against rigid SLIC superpixels.}
\label{fig:supp_ablation_watershed_slic_2}
\end{figure*}

\end{document}

%% file: Algo/Extract_Region_Features.tex
\begin{algorithm}[tb]
\small
\caption{Extract Region Features}
\label{alg:extract_features}
\textbf{Input:} Image $I$ \\
\textbf{Output:} Set of structurally bounded regions $\mathcal{R}$, Pooled region features $\mathcal{F}$
\begin{algorithmic}[1]
\State $M_{fg} \leftarrow \text{ForegroundMask}(I)$ \Comment{Isolate object from background}
\State $F \leftarrow \text{DINOv2}(I)$ \Comment{Extract dense spatial feature map}
\State $\mathcal{R} \leftarrow \text{WatershedPipeline}(I, M_{fg})$ \Comment{Parse into regions}
\State $\mathcal{F} \leftarrow \emptyset$ \Comment{Initialize set for region features}
\For{each region $r \in \mathcal{R}$}
    \State $f_r \gets \frac{1}{|r|} \sum_{p \in r} \ell_2\left(F_p \right)$ \Comment{Average the features $F_p$ of all pixels $p$ inside $r$}
    \State $\mathcal{F} \leftarrow \mathcal{F} \cup \{f_r\}$ \Comment{Add pooled feature to the set}
\EndFor
\Return $\mathcal{R}, \mathcal{F}$
\end{algorithmic}
\end{algorithm}

%% file: Algo/Global_Prototype_Generation.tex
\begin{algorithm}[tb]
\small
\caption{Part-DINO}
\label{alg:part_dino}
\textbf{Input:} Reference set $\mathcal{D}_{ref}$, Test image $I_{test}$, Number of parts $K$, Category set $\mathcal{S}$ \\
\textbf{Output:} Labeled Part Map $\mathcal{L}$
\begin{algorithmic}[1]
\Statex \textbf{\textcolor{blue}{Phase 1: Global Prototype Generation}}
\For{each category $s \in \mathcal{S}$}
    \State $\mathcal{P}^{(s)}, \mathcal{Z}^{(s)} \leftarrow \emptyset, \emptyset$
    \For{each image $I_i \in \mathcal{D}_{ref}^{(s)}$}
        \State $\mathcal{R}_i, \mathcal{F}_i \leftarrow \textsc{Extract Region Features}(I_i)$
        \State $\mathcal{P}^{(s)} \leftarrow \mathcal{P}^{(s)} \cup \mathcal{F}_i$; \quad $\mathcal{Z}^{(s)} \leftarrow \mathcal{Z}^{(s)} \cup \{\text{DINO\_CLS}(I_i)\}$
    \EndFor
    \State $\mathcal{C}^{(s)} \leftarrow \text{K-Means}(\mathcal{P}^{(s)}, K)$ \Comment{$K$ part prototypes}
    \State $\bar{z}_{\text{cls}}^{(s)} \leftarrow \text{mean}(\mathcal{Z}^{(s)})$ \Comment{CLS prototype}
\EndFor

\Statex \textbf{\textcolor{blue}{Phase 2: Region Assignment}}
\State $\mathcal{R}, \mathcal{F} \leftarrow \textsc{Extract Region Features}(I_{test})$
\State $z_{\text{cls}} \leftarrow \text{DINO\_CLS}(I_{test})$
\State $s^* \leftarrow \arg\max_{s \in \mathcal{S}} \left( z_{\text{cls}}^\top \bar{z}_{\text{cls}}^{(s)} \right)$ \Comment{Category routing}
\State $\mathcal{L} \leftarrow \emptyset$
\For{each region $r \in \mathcal{R}$ and feature $\hat{f}_r \in \mathcal{F}$}
    \State $\mathcal{L}(r) \leftarrow \arg\max_{k} \left( \hat{f}_r^\top c_k^{(s^*)} \right)$
\EndFor
\Return $\mathcal{L}$
\end{algorithmic}
\end{algorithm}

%% file: tables/main_cub.tex
\begin{tabular}{lccccccccc}
\toprule
\multirow{2}{*}{Method} & \multicolumn{3}{c}{$K = 4$} & \multicolumn{3}{c}{$K = 8$} & \multicolumn{3}{c}{$K = 16$} \\
\cmidrule(lr){2-4} \cmidrule(lr){5-7} \cmidrule(lr){8-10}
 & NME $\downarrow$ & NMI $\uparrow$ & ARI $\uparrow$ & NME $\downarrow$ & NMI $\uparrow$ & ARI $\uparrow$ & NME $\downarrow$ & NMI $\uparrow$ & ARI $\uparrow$ \\
\midrule
\multicolumn{10}{c}{\textit{Weakly-supervised part discovery}} \\
\midrule
\rowcolor{lightgray!45}\cite{huang2020interpretable} & 11.51 & 29.74 & 14.04 & 11.60 & 35.72 & 15.90 & 12.60 & 43.92 & 21.01 \\
 \cite{pdisconet} & 9.12 & 37.82 & 15.26 & 8.52 & 50.08 & 26.96 & 7.64 & 56.87 & 38.05 \\
\rowcolor{lightgray!45} \cite{choudhury2021unsupervised} & 9.20 & 43.50 & 19.60 & - & 51.50 & 28.30 & - & - & - \\
\cite{pdisconet} + ViT-B/14 & 7.70 & 52.59 & 26.66 & 6.34 & 65.01 & 37.90 & 5.95 & 68.63 & 43.41 \\
\rowcolor{lightgray!45} PDiscoFormer\cite{pdiscoformer} & 8.19 & 52.88 & 23.22 & 6.23 & 67.59 & 41.35 & 6.44 & 69.54 & 49.99 \\
\midrule
\multicolumn{10}{c}{\textit{Unsupervised part discovery}} \\
\midrule
\rowcolor{lightgray!45} \cite{xia2024unsupervised} & 9.92 & 45.22 & 19.40 & 8.58 & 52.42 & 26.42 & 8.27 & 53.52 & 33.95 \\
MPAE \cite{mpae} & 7.75 & 58.93 & 33.73 & 5.53 & 68.24 & 45.85 & 5.14 & 79.00 & 66.56 \\
\midrule
\multicolumn{10}{c}{\textit{Training-Free Part Discovery}} \\
\midrule
\rowcolor{lightgray!45}\cite{amir2021deep} & - & 31.18 & 11.21 & - & 47.21 & 19.76 & - & 50.57 & 26.14 \\
\cite{amir2021deep}$\dagger$ & \underline{9.92}& \underline{39.77} & \underline{17.94} & \underline{6.26}& \underline{54.03} & \underline{27.18} & \underline{4.87}& \underline{56.92} & \underline{35.10} \\

\rowcolor{orange!35} Part-DINO (Ours) & \textbf{8.21} & \textbf{53.93} & \textbf{23.01} & \textbf{6.00} & \textbf{61.84} & \textbf{37.92} & \textbf{4.35} & \textbf{65.45} & \textbf{46.00} \\
\bottomrule
\end{tabular}

%% file: tables/main_celebA.tex
\begin{tabular}{lccccccccc}
\toprule
\multirow{2}{*}{Method} & \multicolumn{3}{c}{$K = 4$} & \multicolumn{3}{c}{$K = 8$} & \multicolumn{3}{c}{$K = 16$} \\
\cmidrule(lr){2-4} \cmidrule(lr){5-7} \cmidrule(lr){8-10}
 & NME $\downarrow$ & NMI $\uparrow$ & ARI $\uparrow$ & NME $\downarrow$ & NMI $\uparrow$ & ARI $\uparrow$ & NME $\downarrow$ & NMI $\uparrow$ & ARI $\uparrow$ \\
\midrule
\multicolumn{10}{c}{\textit{Weakly-supervised part discovery}} \\
\midrule
\rowcolor{lightgray!45}\cite{huang2020interpretable} & 8.75 & 56.69 & 34.74 & 7.96 & 54.80 & 34.74 & 7.62 & 62.22 & 41.01 \\
 \cite{pdisconet} & 11.11 & 75.97 & 69.53 & 9.82 & 62.61 & 51.89 & 9.46 & 77.43 & 70.58 \\
\midrule
\multicolumn{10}{c}{\textit{Unsupervised part discovery}} \\
\midrule
\rowcolor{lightgray!45}GANSeg\cite{he2022ganseg} & 12.26 & 41.71 & 28.06 & 6.18 & 67.28 & 56.23 & - & - & - \\
\cite{xia2024unsupervised} & 12.28 & 50.69 & 34.57 & 10.24 & 58.35 & 46.78 & - & - & - \\
\rowcolor{lightgray!45}MPAE \cite{mpae} & 11.44 & 56.13 & 37.20 & 7.55 & 59.64 & 41.72 & 7.10 & 70.21 & 60.33 \\
\midrule
\multicolumn{10}{c}{\textit{Training-Free part discovery}} \\
\midrule
\rowcolor{lightgray!45}\cite{amir2021deep} & 11.82 & 0.17 & 0.01 & 11.34 & \underline{1.19}& \underline{0.03}& - & \underline{3.29}& \underline{0.06}\\
\cite{amir2021deep} $\dagger$ & \underline{11.36} & \underline{1.38}& \underline{0.01}& \underline{10.74} & 1.12 & 0.01 & \underline{11.42} & 1.99 & 0.03 \\
\rowcolor{orange!35} Part-DINO(Ours) & \textbf{10.81}& \textbf{45.07} & \textbf{32.20} & \textbf{10.59} & \textbf{41.53} & \textbf{28.59} & \textbf{9.15} & \textbf{41.04} & \textbf{25.20} \\
\bottomrule
\end{tabular}

%% file: tables/main_vis.tex
\begin{tabular}{c c@{\hspace{2pt}}c @{\hspace{7pt}} c@{\hspace{2pt}}c @{\hspace{7pt}} c@{\hspace{2pt}}c @{\hspace{7pt}} c@{\hspace{2pt}}c}
    & \multicolumn{2}{c}{\small CUB ($k=16$)} 
    & \multicolumn{2}{c}{\small CelebA ($k=16$)} 
    & \multicolumn{2}{c}{\small PartImage\_S ($k=50$)} 
    & \multicolumn{2}{c}{\small PartImage\_O ($k=50$)} \\
    
    \raisebox{-.5\height}{\rotatebox{90}{\small \text{Image}}} &
    \raisebox{-.5\height}{\includegraphics[width=0.11\linewidth, height=0.11\linewidth]{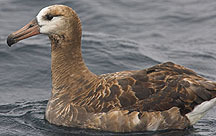}} &
    \raisebox{-.5\height}{\includegraphics[width=0.11\linewidth, height=0.11\linewidth]{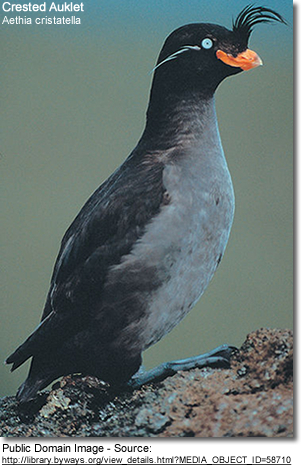}} &
    \raisebox{-.5\height}{\includegraphics[width=0.11\linewidth, height=0.11\linewidth]{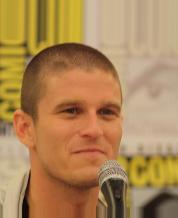}} &
    \raisebox{-.5\height}{\includegraphics[width=0.11\linewidth, height=0.11\linewidth]{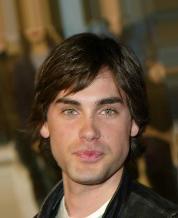}} &
    \raisebox{-.5\height}{\includegraphics[width=0.11\linewidth, height=0.11\linewidth]{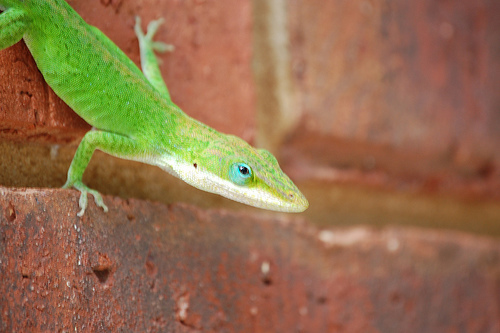}} &
    \raisebox{-.5\height}{\includegraphics[width=0.11\linewidth, height=0.11\linewidth]{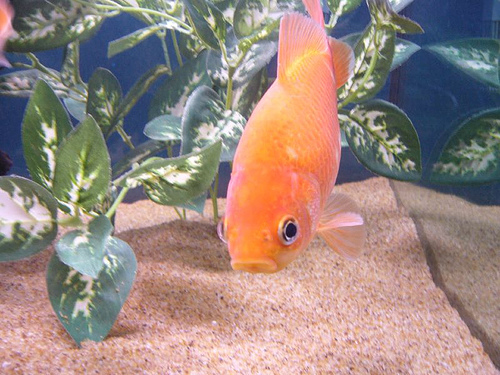}} &
    \raisebox{-.5\height}{\includegraphics[width=0.11\linewidth, height=0.11\linewidth]{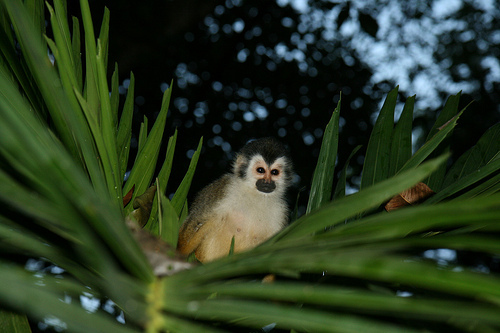}} &
    \raisebox{-.5\height}{\includegraphics[width=0.11\linewidth, height=0.11\linewidth]{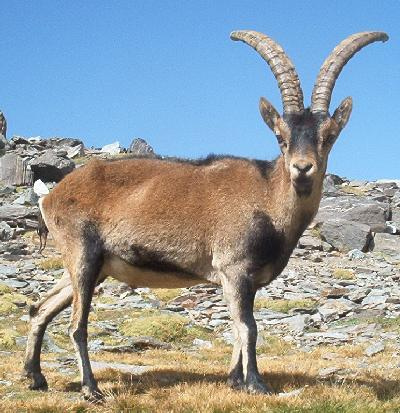}} \\
    
    \raisebox{-.5\height}{\rotatebox{90}{\small \text{MPAE}}} &
    \raisebox{-.5\height}{\includegraphics[width=0.11\linewidth]{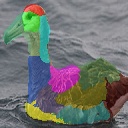}} &
    \raisebox{-.5\height}{\includegraphics[width=0.11\linewidth]{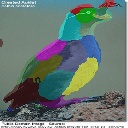}} &
    \raisebox{-.5\height}{\includegraphics[width=0.11\linewidth]{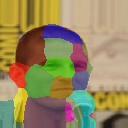}} &
    \raisebox{-.5\height}{\includegraphics[width=0.11\linewidth]{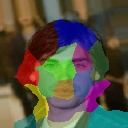}} &
    \raisebox{-.5\height}{\includegraphics[width=0.11\linewidth]{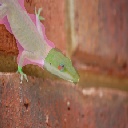}} &
    \raisebox{-.5\height}{\includegraphics[width=0.11\linewidth]{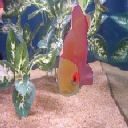}} &
    \raisebox{-.5\height}{\includegraphics[width=0.11\linewidth]{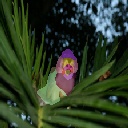}} &
    \raisebox{-.5\height}{\includegraphics[width=0.11\linewidth]{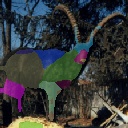}} \\
    
    \raisebox{-.5\height}{\rotatebox{90}{\small \text{Amir}}} &
    \raisebox{-.5\height}{\includegraphics[width=0.11\linewidth]{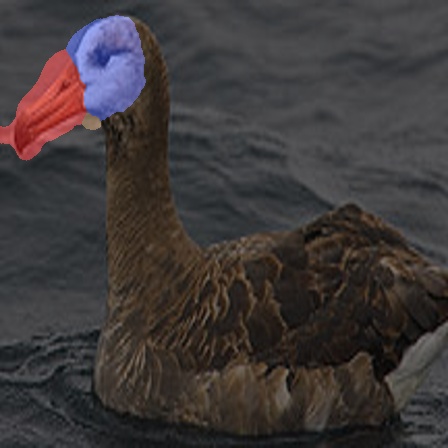}} &
    \raisebox{-.5\height}{\includegraphics[width=0.11\linewidth]{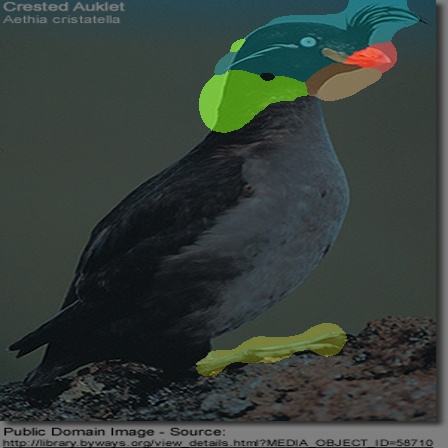}} &
    \raisebox{-.5\height}{\includegraphics[width=0.11\linewidth]{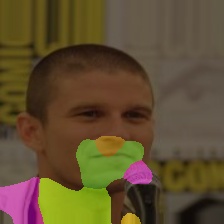}} &
    \raisebox{-.5\height}{\includegraphics[width=0.11\linewidth]{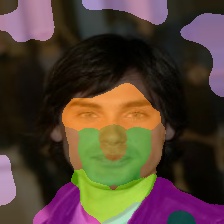}} &
    \raisebox{-.5\height}{\includegraphics[width=0.11\linewidth]{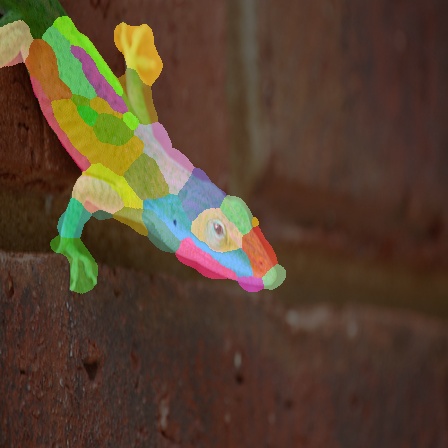}} &
    \raisebox{-.5\height}{\includegraphics[width=0.11\linewidth]{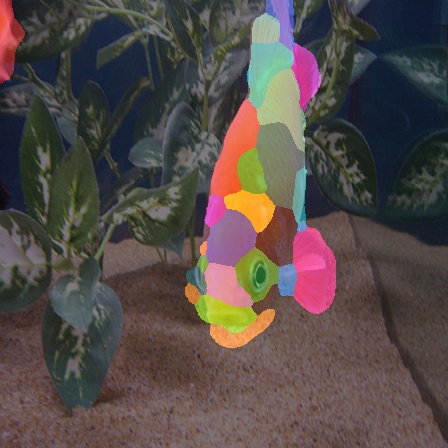}} &
    \raisebox{-.5\height}{\includegraphics[width=0.11\linewidth]{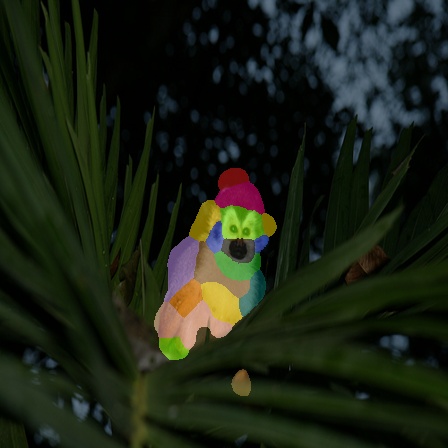}} &
    \raisebox{-.5\height}{\includegraphics[width=0.11\linewidth]{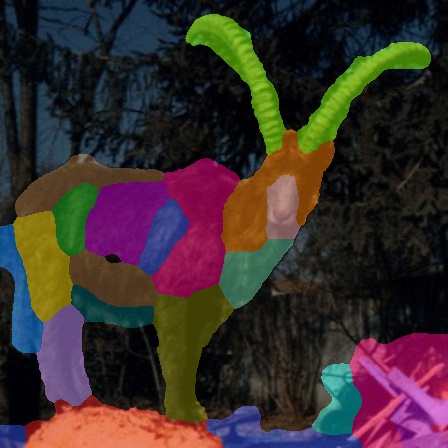}} \\
    
    \raisebox{-.5\height}{\rotatebox{90}{\small \text{Amir +SAM}}} &
    \raisebox{-.5\height}{\includegraphics[width=0.11\linewidth]{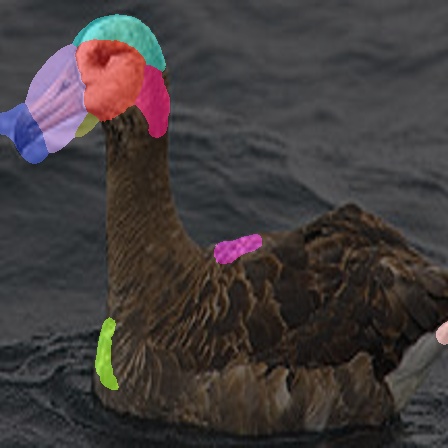}} &
    \raisebox{-.5\height}{\includegraphics[width=0.11\linewidth]{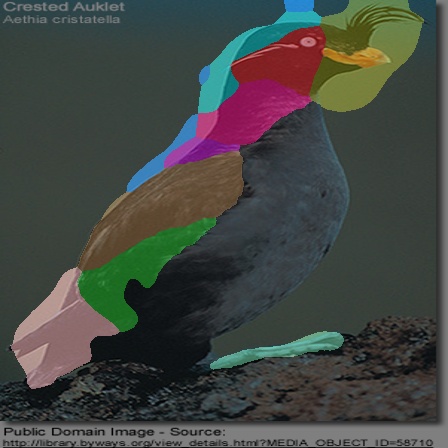}} &
    \raisebox{-.5\height}{\includegraphics[width=0.11\linewidth]{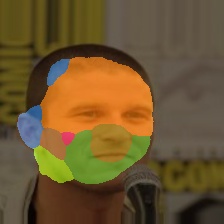}} &
    \raisebox{-.5\height}{\includegraphics[width=0.11\linewidth]{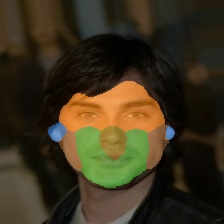}} &
    \raisebox{-.5\height}{\includegraphics[width=0.11\linewidth]{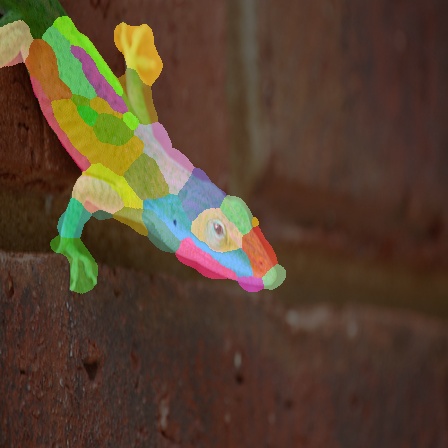}} &
    \raisebox{-.5\height}{\includegraphics[width=0.11\linewidth]{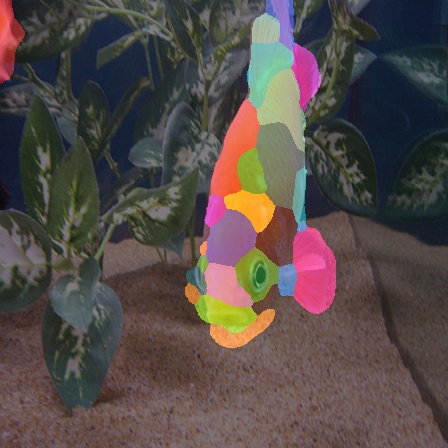}} &
    \raisebox{-.5\height}{\includegraphics[width=0.11\linewidth]{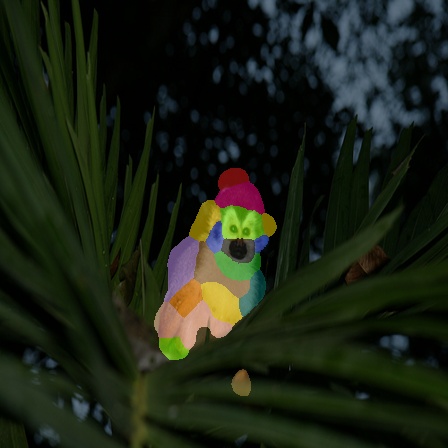}} &
    \raisebox{-.5\height}{\includegraphics[width=0.11\linewidth]{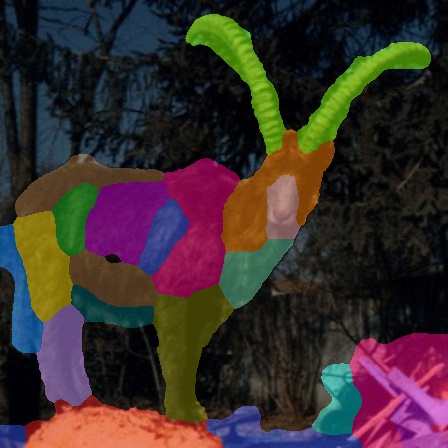}} \\
    
    \raisebox{-.5\height}{\rotatebox{90}{\small \text{Ours}}} &
    \raisebox{-.5\height}{\includegraphics[width=0.11\linewidth]{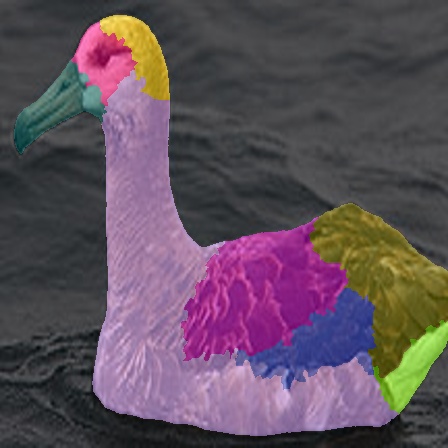}} &
    \raisebox{-.5\height}{\includegraphics[width=0.11\linewidth]{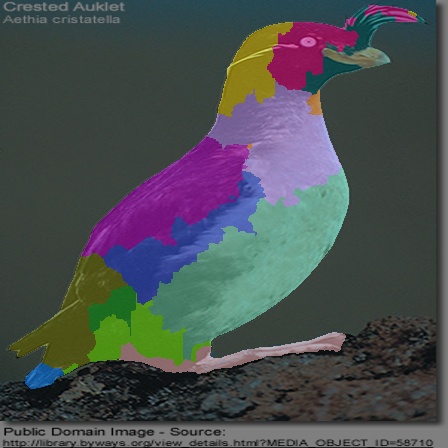}} &
    \raisebox{-.5\height}{\includegraphics[width=0.11\linewidth]{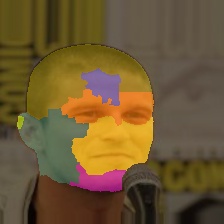}} &
    \raisebox{-.5\height}{\includegraphics[width=0.11\linewidth]{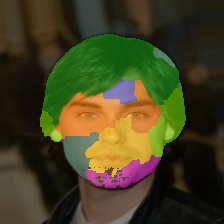}} &
    \raisebox{-.5\height}{\includegraphics[width=0.11\linewidth]{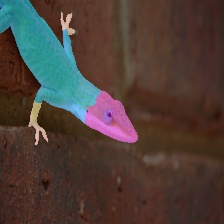}} &
    \raisebox{-.5\height}{\includegraphics[width=0.11\linewidth]{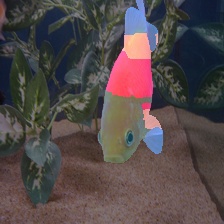}} &
    \raisebox{-.5\height}{\includegraphics[width=0.11\linewidth]{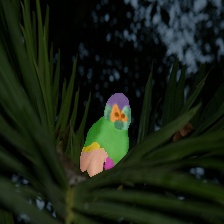}} &
    \raisebox{-.5\height}{\includegraphics[width=0.11\linewidth]{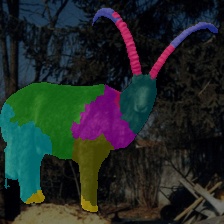}} \\
\end{tabular}

%% file: tables/main_partimagenet.tex
\begin{tabular}{lcccccccccccc}
\toprule
\multirow{3}{*}{Method} & \multicolumn{6}{c}{PartImage$\_O$ (\%)} & \multicolumn{6}{c}{PartImage$\_S$ (\%)} \\
\cmidrule(lr){2-7} \cmidrule(lr){8-13}
 & \multicolumn{2}{c}{$K = 8$} & \multicolumn{2}{c}{$K = 25$} & \multicolumn{2}{c}{$K = 50$} & \multicolumn{2}{c}{$K = 8$} & \multicolumn{2}{c}{$K = 25$} & \multicolumn{2}{c}{$K = 50$} \\
\cmidrule(lr){2-3} \cmidrule(lr){4-5} \cmidrule(lr){6-7} \cmidrule(lr){8-9} \cmidrule(lr){10-11} \cmidrule(lr){12-13}
 & NMI $\uparrow$ & ARI $\uparrow$ & NMI $\uparrow$ & ARI $\uparrow$ & NMI $\uparrow$ & ARI $\uparrow$ & NMI $\uparrow$ & ARI $\uparrow$ & NMI $\uparrow$ & ARI $\uparrow$ & NMI $\uparrow$ & ARI $\uparrow$ \\
\midrule
\multicolumn{13}{c}{\textit{Weakly-supervised part discovery}} \\
\midrule
\rowcolor{lightgray!45}\cite{huang2020interpretable} & 5.88 & 1.53 & 7.56 & 1.25 & 10.19 & 1.05 & - & - & - & - & - & - \\
\cite{pdisconet} & 27.13 & 8.76 & 32.41 & 10.69 & 41.49 & 14.17 & 10.68 & 32.43 & 18.44 & 40.70 & 21.98 & 39.40 \\
\rowcolor{lightgray!45}\cite{pdisconet}(ViT-B/14) & 19.28 & 34.72 & 28.23 & 25.93 & 29.48 & 28.68 & 16.68 & 32.01 & 46.81 & 40.70 & 34.13 & 39.40 \\
\cite{pdiscoformer} (ViT-B/14$^\star$) & 28.84 & 55.66 & 43.36 & 62.82 & 44.48 & 57.91 & - & - & - & - & - & - \\
\rowcolor{lightgray!45}
\cite{pdiscoformer}(ViT-B/14$^\dagger$) & 29.00 & 52.40 & 44.71 & 59.27 & 46.29 & 62.21 & 20.29 & 38.90 & 43.18 & 64.61 & 44.06 & 60.10 \\
\midrule
\multicolumn{13}{c}{\textit{Unsupervised part discovery}} \\
\midrule
\rowcolor{lightgray!45}\cite{xia2024unsupervised} & 27.85 & 40.77 & - & - & - & - & 25.17 & 37.50 & - & - & - & - \\
MPAE\cite{mpae} & 32.90 & 66.17 & 39.28 & 68.87 & 53.65 & 74.22 & 33.51 & 65.05 & 37.22 & 68.04 & 55.10 & 73.52 \\
\midrule
\multicolumn{13}{c}{\textit{Training-Free part discovery}} \\
\midrule
\rowcolor{lightgray!45} \cite{amir2021deep} & 19.17 & 7.59 & \underline{31.46} & 14.16 & \underline{37.81} & 16.50 & - & - & - & - & - & - \\
\cite{amir2021deep} $\dagger$ & \textbf{32.86} &  \underline{54.77}& \textbf{37.19}  &57.87  &\textbf{40.41}  &\underline{59.99}  &25.59  &57.69  & 27.08 &62.92  &32.12  &64.73  \\
\rowcolor{orange!35} Part-DINO(Ours) & \underline{30.07} & \textbf{63.27} & 31.14 & \textbf{64.93} & 30.13 & \textbf{65.42} & \textbf{37.21} & \textbf{66.10} & \textbf{41.54} & \textbf{66.94} & \textbf{41.80} & \textbf{66.97} \\
\bottomrule
\end{tabular}

%% file: checklist.tex
\section*{NeurIPS Paper Checklist}


\begin{enumerate}

\item {\bf Claims}
    \item[] Question: Do the main claims made in the abstract and introduction accurately reflect the paper's contributions and scope?
    \item[] Answer: \answerYes{}.
    \item[] Justification: The abstract and introduction clearly state the contributions made in the paper and important assumptions and limitations. Our experimental results justify our claims

\item {\bf Limitations}
    \item[] Question: Does the paper discuss the limitations of the work performed by the authors?
    \item[] Answer: \answerYes{}.
    \item[] Justification: Yes, we have provided a limitation section in the paper.

\item {\bf Theory assumptions and proofs}
    \item[] Question: For each theoretical result, does the paper provide the full set of assumptions and a complete (and correct) proof?
    \item[] Answer: \answerNA{}.
    \item[] Justification: \answerNA{}.

    \item {\bf Experimental result reproducibility}
    \item[] Question: Does the paper fully disclose all the information needed to reproduce the main experimental results of the paper to the extent that it affects the main claims and/or conclusions of the paper (regardless of whether the code and data are provided or not)?
    \item[] Answer:  \answerYes{}.
    \item[] Justification: See implementation details. 

\item {\bf Open access to data and code}
    \item[] Question: Does the paper provide open access to the data and code, with sufficient instructions to faithfully reproduce the main experimental results, as described in supplemental material?
    \item[] Answer: \answerYes{}.
    \item[] Justification: We provide the code details (implementation details) to faithfully reproduce the main experimental results.

\item {\bf Experimental setting/details}
    \item[] Question: Does the paper specify all the training and test details (e.g., data splits, hyperparameters, how they were chosen, type of optimizer) necessary to understand the results?
    \item[] Answer:  \answerYes{}.
    \item[] Justification: See implementation details section.

\item {\bf Experiment statistical significance}
    \item[] Question: Does the paper report error bars suitably and correctly defined or other appropriate information about the statistical significance of the experiments?
    \item[] Answer: \answerNo{}.
    \item[] Justification: We follow the same experimentation of prior works. Given the use of or centroids, the results will be exactly same.

\item {\bf Experiments compute resources}
    \item[] Question: For each experiment, does the paper provide sufficient information on the computer resources (type of compute workers, memory, time of execution) needed to reproduce the experiments?
    \item[] Answer: \answerYes{}.
    \item[] Justification: The necessary values are presented in the implementation details, results sections and the Appendix.
    
\item {\bf Code of ethics}
    \item[] Question: Does the research conducted in the paper conform, in every respect, with the NeurIPS Code of Ethics \url{https://neurips.cc/public/EthicsGuidelines}?
    \item[] Answer: \answerYes{}
    \item[] Justification: We reviewed the NeurIPS Code of Ethics.584

\item {\bf Broader impacts}
    \item[] Question: Does the paper discuss both potential positive societal impacts and negative societal impacts of the work performed?
    \item[] Answer: \answerYes{}.
    \item[] Justification: Provided in Sec 6. Discussion
    
\item {\bf Safeguards}
    \item[] Question: Does the paper describe safeguards that have been put in place for responsible release of data or models that have a high risk for misuse (e.g., pre-trained language models, image generators, or scraped datasets)?
    \item[] Answer: \answerNA{}.
    \item[] Justification:\answerNA{}.

\item {\bf Licenses for existing assets}
    \item[] Question: Are the creators or original owners of assets (e.g., code, data, models), used in the paper, properly credited and are the license and terms of use explicitly mentioned and properly respected?
    \item[] Answer: \answerYes{}.
    \item[] Justification: We have properly credited and cited the original owners of assets where appropriate.

\item {\bf New assets}
    \item[] Question: Are new assets introduced in the paper well documented and is the documentation provided alongside the assets?
    \item[] Answer:  \answerYes{}.
    \item[] Justification: We provide the necessary details of the code in the paper. We do not have any other assets.

\item {\bf Crowdsourcing and research with human subjects}
    \item[] Question: For crowdsourcing experiments and research with human subjects, does the paper include the full text of instructions given to participants and screenshots, if applicable, as well as details about compensation (if any)? 
    \item[] Answer: \answerNA{}.
    \item[] Justification: \answerNA{}

\item {\bf Institutional review board (IRB) approvals or equivalent for research with human subjects}
    \item[] Question: Does the paper describe potential risks incurred by study participants, whether such risks were disclosed to the subjects, and whether Institutional Review Board (IRB) approvals (or an equivalent approval/review based on the requirements of your country or institution) were obtained?
    \item[] Answer:  \answerNA{}.
    \item[] Justification: \answerNA{}.

\item {\bf Declaration of LLM usage}
    \item[] Question: Does the paper describe the usage of LLMs if it is an important, original, or non-standard component of the core methods in this research? Note that if the LLM is used only for writing, editing, or formatting purposes and does \emph{not} impact the core methodology, scientific rigor, or originality of the research, declaration is not required.
    \item[] Answer: \answerNA{}.
    \item[] Justification: \answerNA{}.

\end{enumerate}